\PassOptionsToPackage{table}{xcolor}
\documentclass{article}
\usepackage[utf8]{inputenc}
\ifdefined\pdfsuppressptexinfo\pdfsuppressptexinfo=-1\fi
\usepackage{iclr2027_conference,times}
\iclrfinalcopy

\usepackage{amsmath,amsfonts,bm}

\def\eqref#1{equation~\ref{#1}}

\def\1{\bm{1}}

\DeclareMathAlphabet{\mathsfit}{\encodingdefault}{\sfdefault}{m}{sl}
\SetMathAlphabet{\mathsfit}{bold}{\encodingdefault}{\sfdefault}{bx}{n}

\usepackage{hyperref}
\usepackage{url}
\usepackage{graphicx}
\usepackage{booktabs}
\usepackage{subcaption}
\usepackage{amsmath}
\usepackage{amssymb}
\usepackage{amsthm}
\usepackage{mathtools}
\usepackage[table]{xcolor}
\usepackage{colortbl}
\usepackage{multirow}
\usepackage{enumitem}
\usepackage{algorithm}
\usepackage{algpseudocode}
\usepackage{threeparttable}

\makeatletter
\let\showhyphens\@showhyphens
\makeatother

\title{Pretraining Shapes Spectral Structure:\\
Architecture- and Strategy-Conditional Prediction of\\
OOD Robustness in Foundation Models}

\author{{Sangyoon Bae} \\
Interdisciplinary Program in Artificial Intelligence \\
Seoul National University\\
Seoul, 08826, South Korea \\
\texttt{stellasybae@snu.ac.kr} \\
\And
Sk Miraj Ahmed \\
Computational Science Initiative \\
Brookhaven National Laboratory \\
Upton, New York, 11973, USA  \\
\texttt{sahmed3@bnl.gov}
\And
Shinjae Yoo \\
Computational Science Initiative \\
Brookhaven National Laboratory \\
Upton, New York, 11973, USA  \\
\texttt{sjyoo@bnl.gov}
\And
Jiook Cha \\
Department of Psychology \\
Seoul National University \\
Seoul, 08826, South Korea  \\
\texttt{connectome@snu.ac.kr}}

\newcommand{\logsr}{\mathrm{log\text{-}sr}}

\newcommand{\logsrf}{\mathrm{log\text{-}sr}_{\mathrm{feat}}}
\newcommand{\logfrob}{\log\|W\|_F}
\newcommand{\logfrobbar}{\overline{\log\|W\|_F}}

\newcommand{\rholoo}{\rho_{\mathrm{LOO}}}

\newcommand{\cvlogfrob}{\mathrm{CV}(\log\text{-Frob})}
\newcommand{\cvpr}{\mathrm{CV}(\mathrm{PR})}
\newcommand{\hentropy}{\bar{H}_{\mathrm{spec}}}
\newcommand{\cvsr}{\mathrm{CV}(\mathrm{SR})}
\newcommand{\prmean}{\mathrm{PR\text{-}mean}}
\newcommand{\prstd}{\mathrm{PR\text{-}std}}
\newcommand{\mpenergy}{\mathrm{MP\text{-}energy}}
\newcommand{\logcond}{\mathrm{log\text{-}cond}}

\newcommand{\lognfr}{\mathrm{log\text{-}NFR}}

\newcommand{\entropymean}{\overline{H}_{\mathrm{ent}}}
\newcommand{\alphamean}{\bar{\alpha}}

\providecommand{\sr}{\mathrm{sr}}

\providecommand{\Dacc}{\Delta_{\mathrm{acc}}}

\providecommand{\Fsrc}{F_{\mathrm{src}}}

\newtheorem{theorem}{Theorem}
\newtheorem{lemma}[theorem]{Lemma}
\newtheorem{proposition}[theorem]{Proposition}

\definecolor{clmbg}{HTML}{D6E4FF}
\definecolor{mlmbg}{HTML}{D0EDE6}
\definecolor{ctvbg}{HTML}{FFF0D5}
\definecolor{supbg}{HTML}{EAE0FF}
\definecolor{invbg}{HTML}{FFD6D6}
\definecolor{mrgbg}{HTML}{FFE8CC}


\begin{document}

\maketitle

\begin{abstract}
Can we determine whether a foundation model will generalize out-of-distribution (OOD) before any target data is available?
Existing diagnostics require source or target data, which rules them out before a target domain exists.
Those that use the weights alone apply one statistic to every architecture, and do not separate robust models from fragile ones.
We show the answer is encoded in the spectral structure of pretrained weights.
Two forces shape that structure.
Architecture determines \emph{how} information is stored in weight matrices.
Pretraining strategy determines \emph{what} is rewarded.
Together they set a spectral geometry that governs OOD robustness.
We prove that the OOD accuracy gap is bounded by how tightly the source representations concentrate.
A statistic computed from the pretrained weights alone serves as a proxy for that concentration.
The direction of that proxy reverses between architecture families.
We operationalize it: the direction is stable within one (architecture $\times$ strategy) combination, the finest grouping we test, which we call a cell.
Pooled over 116 models spanning 7 modalities, a single statistic ranks OOD robustness weakly, because cells of opposite direction cancel.
Within a cell, the statistic selected for it orders $92\%$ of model pairs by OOD robustness in-sample.
The selection does not leak the target: for each model family outside the matrix we logged the cell, metric and sign before running its OOD evaluation, and the predicted direction held in every case: EEG, genomic and protein.
Acting on spectral concentration narrows the OOD gap by $24\%$ at $87.5\%$ ID retention.
The diagnostic operates on released weights alone, so OOD robustness becomes checkable at model-selection time, before data or compute is committed to a target domain.
\end{abstract}

\section{Introduction}
\label{sec:intro}

Can we determine whether a foundation model will generalize out-of-distribution \emph{before any target data is available?}
Data-dependent diagnostics~\citep{sun2022out,liu2020energy,yu2025odp} require labeled source examples or target access, ruling them out at release time.
Weight-space approaches~\citep{martin2021predicting,sanyal2019stable} apply one statistic to every architecture, and rank models at correlations around $0.4$--$0.5$.
That ceiling has been attributed to the noisiness of spectral signals.
We argue the cause is the population, not the statistic.
Pooling models whose spectra were shaped by different forces flattens the signal~\citep{martin2021post}.

\begin{figure}[!ht]
\centering
\includegraphics[width=0.82\linewidth, height=0.82\textheight, keepaspectratio]{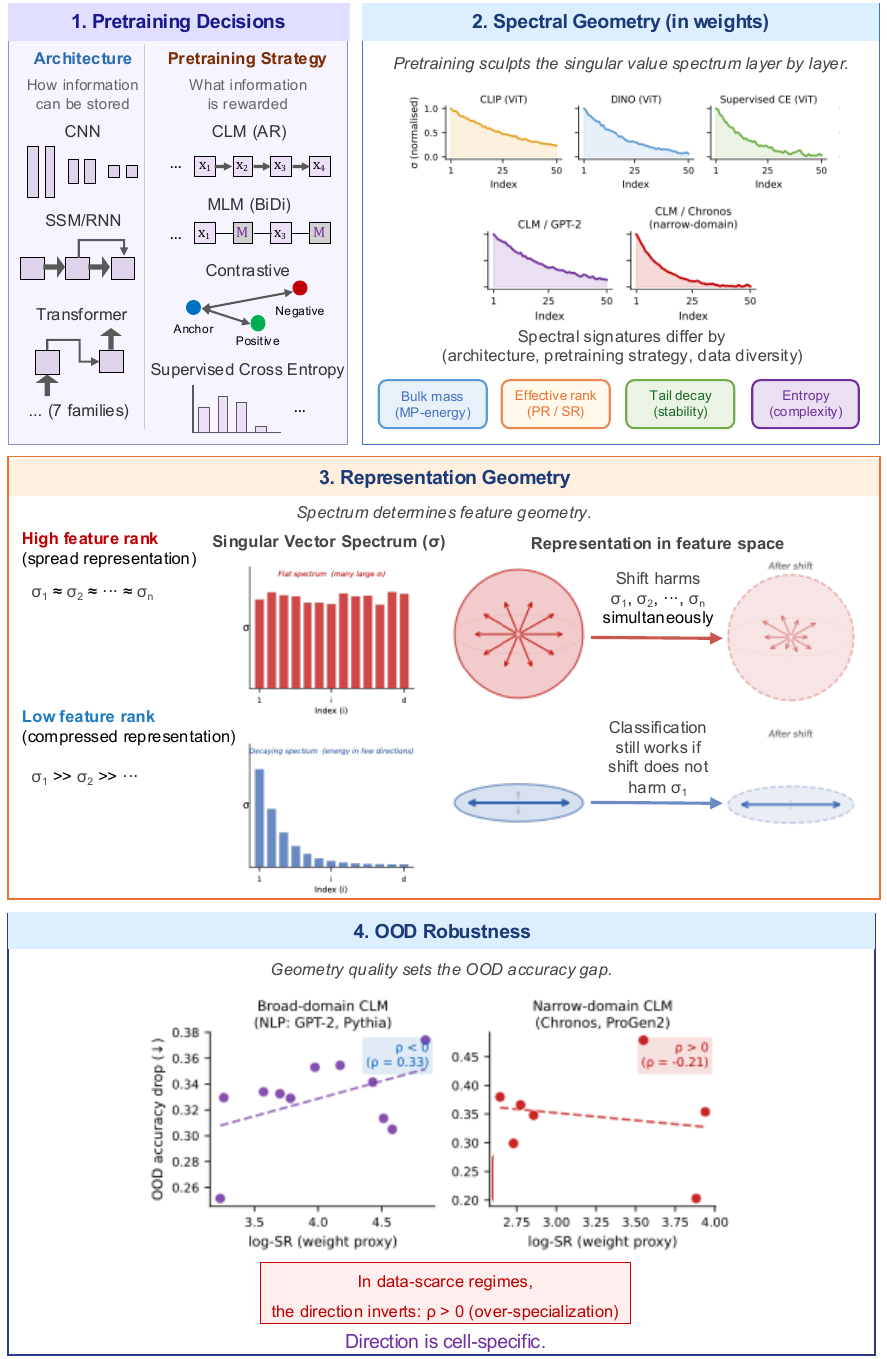}
\caption{
  \textbf{(Architecture $\times$ Strategy) $\to$ Spectral Geometry $\to$ OOD Robustness.}
  Architecture and pretraining strategy jointly shape the layer-wise singular-value spectrum \textbf{(2)}, which determines whether representations are concentrated or diffuse \textbf{(3)}, setting the OOD accuracy gap \textbf{(4)}.
  The sign of $\rho$ is a property of the cell and of the statistic read in it, not of the domain: each AR-CLM cell is positive under its own calibrated metric, while BiDi-MLM/Protein, CNN and SSM-v2/RWKV are negative under theirs.
}
\label{fig:matrix}
\end{figure}

Each (architecture $\times$ strategy) combination leaves a distinct geometric fingerprint in pretrained weights---because architecture controls \emph{how} information is stored and pretraining strategy controls \emph{what} is rewarded---that determines how representations behave under distribution shift (Figure~\ref{fig:matrix}).
Pooling across incompatible fingerprints cancels the OOD-predictive signal; conditioning on them restores it.

\textbf{Contributions.}
\begin{itemize}[nosep,leftmargin=*]
\item \textbf{A spectral-concentration bound explains why conditioning must reach (architecture $\times$ strategy), the finest grouping we test.} Pooling across cells retains under a third of the correlation that per-cell conditioning achieves, and every one of the 15 cells achieves $|\rholoo| \geq 0.700$ (Theorem~\ref{thm:main}, Table~\ref{tab:hierarchy}).
\item \textbf{Acting on spectral concentration narrows the OOD gap by 24\% at 87.5\% ID retention.} Metric-guided layer selection beats random targeting in 9 of 10 models.
\item \textbf{The diagnostic transfers data-free to model families outside the matrix.} The cell, metric and sign were logged before each new family was evaluated, and the predicted direction held in the EEG, genomic and protein cases.
\end{itemize}

\section{Related Work}
\label{sec:related}

Data-free approaches tie spectral statistics of pretrained weights to model quality~\citep{martin2021predicting,sanyal2019stable,neyshabur2018pac}, and \citet{martin2021post} report that this relationship can reverse sign between architectural subgroups---a prior finding that we operationalize as a 15-cell (architecture $\times$ strategy) registry with a statistic selected for each cell, extended from ID generalization to OOD robustness (Appendix~\ref{app:extended_related}).

\paragraph{Where the spectral geometry comes from.}
OOD robustness is governed by representation geometry: the Tunnel Effect~\citep{masarczyk2023tunnel} links OOD linear-probe accuracy to feature rank, Domain Feature Collapse~\citep{yang2025domainfeature} shows single-domain pretraining reduces feature dimensionality, and \citet{chou2026diagnosing} and \citet{zia2026geometry} extend this to geometric measures of representation space---all of them requiring source-domain access.
That geometry is in turn shaped by the architecture and the pretraining objective, which impose characteristic pressures: compact low-dimensional representations under contrastive training, stage-wise specialization in hierarchical backbones, depth-structured residuals in autoregressive Transformers, concentrated spectral mass under state-space recurrence~\citep{radford2021learning,caron2021emerging,liu2021swin,elhage2021mathematical,clark2019does,jing2021understanding,garrido2023rankme,gu2022efficiently,gu2023mamba}.
We read one spectral statistic per (architecture $\times$ strategy) cell as a weight-space proxy for the same quantities---motivated by those pressures, though the final choice in each cell was made empirically (Appendix~\ref{app:models}); Appendix~\ref{app:extended_related} gives the per-architecture and per-objective accounts in full.

\section{Theory: Spectral Concentration Predicts OOD Robustness}
\label{sec:theory}

Prior work links representation rank and geometry to OOD behavior~\citep{masarczyk2023tunnel,yang2025domainfeature}. We formalize this relationship within architecture--strategy cells and then derive weight-only spectral proxies. Complete assumptions, notation, and proofs are deferred to Appendix~\ref{app:proof}.

\begin{theorem}[Log stable rank $\to$ OOD accuracy gap]
\label{thm:main}
For source features $F,F'$ from two checkpoints within the same architecture--strategy cell, let
$\sr(F)=\|F\|_F^2/\|F\|_2^2$. Under T1--T6 (Appendix~\ref{app:proof:main}),
\begin{equation}
\boxed{
\logsr(F')>\logsr(F)
\;\Longrightarrow\;
\Dacc(F')-\Dacc(F)
\ge (1-\delta)q-\delta>0,
}
\label{eq:main-ordinal-bound}
\end{equation}
where $q>0$ is the induced target-error separation and $\delta$ is the joint concentration failure probability. Feature log stable rank orders OOD degradation within the regime characterized by T1--T6.
\end{theorem}

\noindent\textit{Intuition.}
Spreading spectral mass into weaker feature directions exposes more directions to target shift:
\[
\text{weak-tail spreading}
\;\Rightarrow\;
\logsrf\uparrow
\;\Rightarrow\;
\text{target-margin stability}\downarrow
\;\Rightarrow\;
\Dacc\uparrow.
\]

\begin{lemma}[Nested weight-to-feature transfer]
\label{lem:gram-feat}
For a depth-$L$ linear residual model
$X_{\ell+1}=X_\ell(I+W_\ell)$, let
\[
F=X_0M,
\qquad
M=\prod_{\ell=1}^{L}(I+W_\ell).
\]
Under approximate input isotropy,
\begin{subequations}\label{eq:nested-pair}
\begin{gather}
\boxed{|\logsr(F)-\logsr(M)|\le E_X}
\label{eq:nested-global}\\[1mm]
\boxed{|\logsr(F)-\logsr(I+W_\ell)|\le E_X+D_\ell}
\label{eq:layer-transfer}
\end{gather}
\end{subequations}
where
\[
E_X=\log\frac{1+\epsilon_X}{1-\epsilon_X},
\qquad
D_\ell=2\log\kappa(P_\ell)+2\log\kappa(Q_\ell),
\]
for $M=P_\ell(I+W_\ell)Q_\ell$. Hence
$\ell^\star=\arg\min_\ell D_\ell$ gives the tightest certified layerwise approximation. Residual-stream analyses motivate this operator view~\citep{elhage2021mathematical,zangrando2024neural}; the nonlinear analogue follows locally through the block Jacobian. Proof: Appendix~\ref{app:proof:lemma2}.
\end{lemma}

\noindent\textit{Intuition.}
The complete residual product, rather than an isolated layer, determines the final feature geometry:
\[
\{W_\ell\}_{\ell=1}^{L}
\;\Rightarrow\;
M
\;\Rightarrow\;
\logsrf
\;\xRightarrow{\text{Thm.~\ref{thm:main}}}\;
\Dacc.
\]

All candidate statistics below are computed directly from pretrained weights via WeightWatcher~\citep{martin2021predicting}; metric definitions are given in Appendix~\ref{app:mechanistic}.

\begin{proposition}[$\cvlogfrob$ for AR-CLM]
\label{prop:cvlogfrob}
Let $z_\ell=\log\|W_\ell\|_F$ and
$\ell_F=\arg\max_\ell z_\ell$. AR-CLM depth structure~\citep{elhage2021mathematical,fernando2026dynamics} and heavy-tailed self-regularization~\citep{martin2021predicting,everett2026transforming} motivate P3.1--P3.4: dominant variation through $z_{\ell_F}$, decreasing fitted tail exponent $\hat\alpha$ along that checkpoint direction, a common power-law spectral range, and an effective-block gap larger than the distortion in Lemma~\ref{lem:gram-feat}. Then
\begin{equation}
\boxed{
\cvlogfrob\uparrow
\;\Rightarrow\;
\hat\alpha_{\ell_F}\downarrow
\;\Rightarrow\;
\logsr(I+W_{\ell_F})\uparrow
\;\Rightarrow\;
\logsrf\uparrow
\;\Rightarrow\;
\Dacc\uparrow.
}
\label{eq:prop3-main}
\end{equation}
Proof: Appendix~\ref{app:proof:prop3}.
\end{proposition}

\noindent\textit{Intuition.}
A stronger depth-wise norm hierarchy is associated with heavier spectral tails in the dominant AR-CLM block, which increases the effective dimensionality transmitted to the representation.

\begin{proposition}[$\prmean$ polarity in BiDi-MLM]
\label{prop:cvpr}
BiDi-MLM heads develop complementary specialization~\citep{clark2019does}, while the participation ratio gives a scale-consistent measure of effective rank~\citep{jha2025spectral}. Under P4.1--P4.4 and
\[
\Sigma_{\rm feat}
\simeq
\frac{1}{L}
\sum_{\ell}
W_\ell^{OV}(W_\ell^{OV})^\top,
\]
\begin{equation}
\boxed{
\sr_{\rm feat}
=
\frac{\sqrt L}
{\kappa_\Sigma\sqrt{H_P^{-1}+A}},
\qquad
s_{\mathcal C}
=
\operatorname{sign}\!\left(
\frac{d\logsrf}{d\prmean}
\right),
}
\label{eq:pr-core}
\end{equation}
where $H_P$ is the harmonic mean of layerwise participation ratios, $A$ measures cross-layer alignment, and $\kappa_\Sigma$ measures concentration of the aggregate covariance. Approximately stable $A$ and $\kappa_\Sigma$ give $s_{\mathcal C}=+1$, whereas sufficiently strong changes in alignment or aggregate concentration can produce $s_{\mathcal C}=-1$. Proof: Appendix~\ref{app:proof:prop4}.
\end{proposition}

\noindent\textit{Intuition.}
Participation ratio controls within-layer spectral spreading, while $A$ and $\kappa_\Sigma$ determine how those layerwise directions combine:
\[
\prmean\uparrow
\;\Rightarrow\;
(H_P,A,\kappa_\Sigma)
\;\Rightarrow\;
\begin{cases}
\logsrf\uparrow \Rightarrow \Dacc\uparrow, & s_{\mathcal C}=+1,\\[2mm]
\logsrf\downarrow \Rightarrow \Dacc\downarrow, & s_{\mathcal C}=-1.
\end{cases}
\]
with the sign of the first mapping determined by $s_{\mathcal C}$.

\paragraph{What the reported statistics measure.}
Two names need pinning down, because the quantity we report is not always the one the
derivation is written in.
$\prmean$ is the mean stable rank $\|W\|_F^2/\|W\|_2^2$ over layers, and $\cvpr$ is its
coefficient of variation; $\cvsr$ is the coefficient of variation of $\log$ stable rank.
Stable rank is the quantity Theorem~\ref{thm:main} bounds, so the measurement sits on the
same axis as the theorem.
Proposition~\ref{prop:cvpr} above is derived for the participation ratio, which is a
different functional of the same spectrum, and we do not claim the derivation transfers to
the stable rank unchanged.
$\entropymean$ is the entropy of the distribution of fitted power-law exponents across
layers, not of the singular-value spectrum, and $\mpenergy$ is the mean Marchenko--Pastur
soft rank over layers.

\begin{proposition}[MP-energy for ViT$\times$CLIP]
\label{prop:mpclip}
Contrastive objectives induce low-rank pressure~\citep{jing2021understanding}. If InfoNCE concentrates $k$ directions above the Marchenko--Pastur bulk and
$\mathrm{SR}(W)\approx k/(1-\mathrm{MP}(W))$, then
\[
\boxed{
\mpenergy\uparrow
\;\Rightarrow\;
\mathrm{SR}(W)\uparrow \;\Rightarrow\; \logsrf\uparrow \;\Rightarrow\; \Dacc\uparrow.
}
\]
Proof: Appendix~\ref{app:proof:prop5}.
\end{proposition}

\noindent\textit{Intuition.}
Greater bulk energy spreads mass into more directions, yielding a less concentrated representation and a larger OOD accuracy gap.

\paragraph{Cell-specific polarity.}
\label{sec:axes}
Theorem~\ref{thm:main} fixes the feature-level relation between $\logsrf$ and OOD degradation; architecture, pretraining strategy, and data diversity determine which weight statistic tracks that geometry and with what polarity~\citep{ba2024data}. In particular, BiDi-MLM/Protein ($\rholoo=-0.891$) corresponds to an inverted $\prmean\!\to\!\logsrf$ mapping under Proposition~\ref{prop:cvpr}. The resulting within-cell hierarchy is
\[
\boxed{
\text{weight-space proxy}
\;\xRightarrow{\text{Props.~\ref{prop:cvlogfrob}--\ref{prop:mpclip}}}\;
\logsrf
\;\xRightarrow{\text{Thm.~\ref{thm:main}}}\;
\Dacc,
}
\]
with Lemma~\ref{lem:gram-feat} providing the weight-to-feature bridge.

\section{Testing the Theory}
\label{sec:testing}

Does the predicted relationship emerge across foundation models?
We test this across 116 foundation models spanning 7 modalities, each cell on one canonical OOD shift (Appendix~\ref{app:ood_eval}); Appendix~\ref{app:ood-type} stratifies the same data by shift type.

\subsection{Within Every Cell, a Spectral Statistic Orders Models by OOD Robustness}
\label{sec:matrix}

Table~\ref{tab:matrix} lists, for each primary (architecture $\times$ strategy) cell, the statistic selected for it and that statistic's correlation with OOD drop.
Every cell reaches $|\rholoo| \geq 0.700$ (9/15 cells two-sided $p \leq 0.05$, 4 more one-sided under the pre-specified direction, 2 marginal; mean $0.824$; per-cell intervals in Appendix~\ref{app:bootstrap_ci}).

\begin{table}[t]
\centering
\small
\caption{
  \textbf{Architecture $\times$ strategy matrix: numerical summary.}
  Stars: $^{***}$ $|\rholoo| \geq 0.7$, two-sided $p \leq 0.05$; $^{*}$ one-sided $p \leq 0.05$ (pre-specified direction); $^\dagger$ marginal ($|\rholoo| \geq 0.8$, $p > 0.05$ one-sided).
  For $n{=}5$: exact permutation; $n{=}4$: exact one-sided $p{=}0.042$; otherwise Fisher-$z$.
  $\uparrow$: higher metric $\to$ higher OOD drop.
  Row shading encodes correlation sign: salmon ($\rholoo{>}0$, higher metric $\to$ more fragile); lavender ($\rholoo{<}0$, capacity or gating inversion); green (anti-collapse MLM Protein); orange (narrow-domain genomic MLM).
  EEG foundation model results are reported separately in \S\ref{sec:crossdomain}.
}
\label{tab:matrix}
\setlength{\tabcolsep}{4pt}
\begin{tabular}{llllrr}
\toprule
Architecture & Example models & Strategy & Metric & $\rholoo$ & $n$ \\
\midrule
\rowcolor{invbg}
AR-Transformer   & GPT-2 / Pythia / GPT-Neo & CLM           & $\logfrobbar\;\uparrow$  & $+0.714^{***}$ & 16 \\
\rowcolor{invbg}
AR-Transformer   & Chronos / Timer / Sundial & CLM          & $\prmean\;\uparrow$      & $+0.714^{*}$   &  8 \\
\rowcolor{invbg}
AR-Transformer   & ProGen2 / ProtGPT2 / RITA & CLM          & $\cvlogfrob\;\uparrow$   & $+0.817^{***}$ & 10 \\
\rowcolor{invbg}
BiDi-Transformer & BERT / RoBERTa / ALBERT  & MLM           & $\prmean\;\uparrow$      & $+0.808^{***}$ & 10 \\
\rowcolor{mlmbg}
BiDi-Transformer & ESM / AnKH / AMPLIFY     & MLM           & $\prmean$                & $-0.891^{***}$ & 11 \\
\rowcolor{mrgbg}
BiDi-Transformer & NT-v2 / AgroNT (Genomic) & MLM           & $\logfrobbar\;\uparrow$  & $+0.829^{***}$ &  7 \\
\rowcolor{invbg}
ViT              & CLIP-ViT / SigLIP        & Contrastive   & $\mpenergy\;\uparrow$    & $+0.750^{*}$   &  6 \\
\rowcolor{invbg}
ViT              & DINOv2 / DINO            & Contrastive   & $\mpenergy\;\uparrow$    & $+0.800^{\dagger}$ &  5 \\
\rowcolor{invbg}
ViT / DeiT       & ViT-T/S/B/L, DeiT-B/S   & Supervised CE & $\mpenergy\;\uparrow$    & $+0.714^{*}$   &  7 \\
\rowcolor{invbg}
Swin             & Swin/SwinV2-T/S/B/L      & Supervised CE & $\mpenergy\;\uparrow$    & $+0.893^{***}$ &  8 \\
\rowcolor{supbg}
CNN              & ResNet / ConvNeXt        & Supervised CE & $\cvsr$                  & $-0.829^{*}$   &  7 \\
\rowcolor{invbg}
Conv+Transformer & wav2vec2 (XLSR series)   & Contrastive   & $\cvlogfrob\;\uparrow$   & $+0.800^{***}$ &  6 \\
\rowcolor{invbg}
SSM (Mamba)         & Mamba-130M--2.8B      & CLM NLP       & $\entropymean\;\uparrow$ & $+0.800^{\dagger}$ &  5 \\
\rowcolor{supbg}
SSM-v2 (Mamba-2)    & Mamba2-130M--2.7B     & CLM NLP       & $\alphamean$             & $-1.000^{***}$  &  5 \\
\rowcolor{supbg}
RWKV                & RWKV-4-169M--7B       & CLM NLP       & $\cvlogfrob$             & $-1.000^{***}$  &  5 \\
\bottomrule
\end{tabular}
\end{table}

Three cell-level patterns confirm the theory-predicted structure.
First, neither axis alone fixes the metric: holding the architecture and varying the strategy, or the reverse, changes which statistic the theory admits as a candidate and which one calibration then selects (Table~\ref{tab:patterns}).
Second, the direction is cell-specific rather than universal---positive in the salmon- and orange-shaded cells of Table~\ref{tab:matrix}, negative in the BiDi-MLM/Protein, CNN, and SSM-v2/RWKV cells.
Third, per-cell selection achieves strong signal in every case ($|\rholoo| \in [0.714, 1.000]$; vision cells $[0.714, 0.893]$).

\begin{table}[ht]
\centering\small
\caption{
  \textbf{The cell fixes the metric; neither axis alone does.}
  Holding one axis and varying the other changes the statistic selected for the cell.
  Each row reads off Table~\ref{tab:matrix}.
}
\label{tab:patterns}
\setlength{\tabcolsep}{5pt}
\begin{tabular}{lll}
\toprule
Held fixed & Varied & Theory-named metric \\
\midrule
BiDi-Transformer & MLM-NLP $\to$ MLM-Genomic          & $\prmean \to \logfrobbar$ \\
Supervised CE    & ViT / Swin $\to$ CNN               & $\mpenergy \to \cvsr$ \\
CLM              & NLP $\to$ time-series $\to$ protein & $\logfrobbar \to \prmean \to \cvlogfrob$ \\
CLM-NLP          & Mamba $\to$ Mamba-2 $\to$ RWKV     & $\entropymean \to \alphamean \to \cvlogfrob$ \\
\bottomrule
\end{tabular}
\end{table}
Conditioning on (architecture $\times$ strategy) raises mean $|\rholoo|$ from $0.245$ (global pooled, Table~\ref{tab:hierarchy} Level 0) to $0.824$; substituting $\log n_\mathrm{params}$ at the same level drops it to $0.686$, but not in every cell: $\log n_\mathrm{params}$ ties or beats the selected metric in the ViT$\times$CLIP, ViT$\times$Supervised, CNN, Mamba, Mamba-2 and RWKV cells (Appendix~\ref{app:param}).
LOCO stability: dropping any single cell leaves mean $|\rholoo| \in [0.811, 0.832]$ (s.d.\ $0.006$ across 15 rounds, max deviation from baseline $0.013$; Appendix~\ref{app:loco}).
Across 454 within-cell pairwise comparisons, the cell-specific statistic identifies the more OOD-robust model in \textbf{92\%} of cases (pairwise ranking accuracy $0.924$, median $0.958$).

\subsection{Architecture \texorpdfstring{$\times$}{x} Strategy Is the Finest Grouping Still Needed}
\label{sec:conditioning}

Minimality is tested from both sides: coarser groupings below, and attempts to widen a cell by pooling related objectives in Appendix~\ref{app:expansion}.

\begin{table}[ht]
\centering\small
\caption{
  Four-level conditioning hierarchy on the \emph{identical} $n{=}116$ model set.
  Mean $|\rholoo|$ across groups at each level; best metric selected independently
  per group (oracle upper-bound at each level).
  Neither axis alone attains the full cell-level signal.
}
\label{tab:hierarchy}
\setlength{\tabcolsep}{5pt}
\begin{tabular}{clrr}
\toprule
Level & Conditioning unit & Mean $|\rholoo|$ & $\Delta$ \\
\midrule
0 & Global (all 116 models pooled)                       & 0.245 & ---      \\
1 & Strategy-only (4 groups; one metric per objective)   & 0.478 & $+0.233$ \\
2 & Architecture-only (8 groups; one metric per backbone)& 0.745 & $+0.267$ \\
3 & \textbf{Architecture $\times$ Strategy} (ours)       & \textbf{0.824} & $\mathbf{+0.579}$ \\
\rowcolor{gray!12}
3$'$ & Arch $\times$ Strategy (\emph{$\log n_\mathrm{params}$ instead}) & 0.686 & $+0.441$ \\
\bottomrule
\end{tabular}
\smallskip\par\noindent{\footnotesize
  All rows use the identical $n{=}116$ model set; Levels 0--2 use oracle best metric per group (upper-bound at each level).
  Replacing the cell-specific spectral metric with $\log n_\mathrm{params}$ (row 3$'$) drops mean $|\rholoo|$ by $0.138$ (0.824~$\to$~0.686)---below the architecture-only baseline.
}
\end{table}

Table~\ref{tab:hierarchy} and Appendix~\ref{app:conditioning} show a clean monotone hierarchy on the identical $n{=}116$ set: each additional conditioning axis adds signal ($+0.233$ for strategy, $+0.267$ for architecture, $+0.079$ for the interaction), with a total gain of $+0.579$ over the unconditioned baseline.
Neither axis is individually sufficient: strategy-only ($0.478$) and architecture-only ($0.745$) both fall well below the cell level ($0.824$), because incompatible metric directions appear \emph{within} strategy classes (BiDi-MLM Protein uses $\prmean$ at $\rholoo{=}{-}0.891$; BiDi-MLM Genomic uses $\logfrobbar$ at $\rholoo{=}{+}0.829$---mixing them suppresses signal) and \emph{within} architecture classes (ViT spans CLIP/DINO/Supervised with incompatible OOD benchmarks; AR-Transformer spans NLP/protein/time-series with different metrics).
Cell-specific statistics outperform unconditioned alternatives: $\alpha_\mathrm{mean}$ (WeightWatcher) achieves mean $|\rholoo| = 0.448$; the best fixed weight-space metric ($\overline{\rm PR}$) achieves $0.724$; $\log n_\mathrm{params}$ achieves $0.686$---all below the cell-conditioned mean of $0.824$ (parameter count versus weight geometry in Appendix~\ref{app:param}; architecture-conditional composites in Appendix~\ref{app:composites}).
Data-dependent baselines (Doctor score, Mahalanobis distance) reach mean absolute LOO correlations of $0.694$ and $0.562$ in the six and four cells where they were evaluated but require a source-domain test set and cannot transfer across modalities or be applied data-free (Appendix~\ref{app:data-dep}).

\subsection{Training Data Diversity Can Change the Proxy Polarity}
\label{sec:inversion}

Architecture and pretraining strategy determine \emph{which} metric is diagnostic.
Training data diversity can change the \emph{proxy polarity}: narrow-domain or data-scarce pretraining can reverse the observed weight-metric--OOD direction while leaving the feature-level direction in Theorem~\ref{thm:main} unchanged.
We classify pretraining as \emph{broad-domain} when the corpus spans diverse topics, species, or registers at $\gtrsim 10^8$ tokens (e.g., The Pile for NLP, UniRef90 for proteins); \emph{narrow-domain} when constrained by limited vocabulary (4-nucleotide or 20-AA alphabets), single task domain, or limited data volume ($\lesssim 10^6$ sequences).

Within-family contrasts confirm this: holding (architecture $\times$ strategy) fixed and varying only the training domain shifts the selected metric while the sign is preserved within each family that shares a pretraining objective (Appendix~\ref{app:scarcity}).
Across broad- and narrow-domain pretraining alike, the cell-specific metric reliably predicts OOD robustness ($|\rholoo|$ ranges from $0.714$ to $1.000$ across all 15 cells).
The SSM/RWKV cells show high $|\rholoo|$ with different metrics than Transformer cells ($\entropymean$ for Mamba, $\alphamean$ for Mamba-2, $\cvlogfrob$ for RWKV), reflecting distinct weight-space geometry from selective gating (Appendices~\ref{app:mechanistic} and~\ref{sec:ssm}).
The CNN cell is the only broad-domain cell where $\rholoo < 0$ with $\cvsr$, consistent with a capacity-scaling regime where larger models are more robust; Swin and ViT cells also use $\mpenergy$ but show $\rholoo > 0$, indicating distinct geometric regimes across vision architectures.
The CLM objective shapes the \emph{type} of spectral geometry across modalities: $\logfrobbar$ in NLP, $\prmean$ in time-series, $\cvlogfrob$ in protein.

\section{Acting on Spectral Concentration Narrows the OOD Gap}
\label{sec:intervention}

\begin{figure}[t]
\centering
\includegraphics[width=0.97\linewidth]{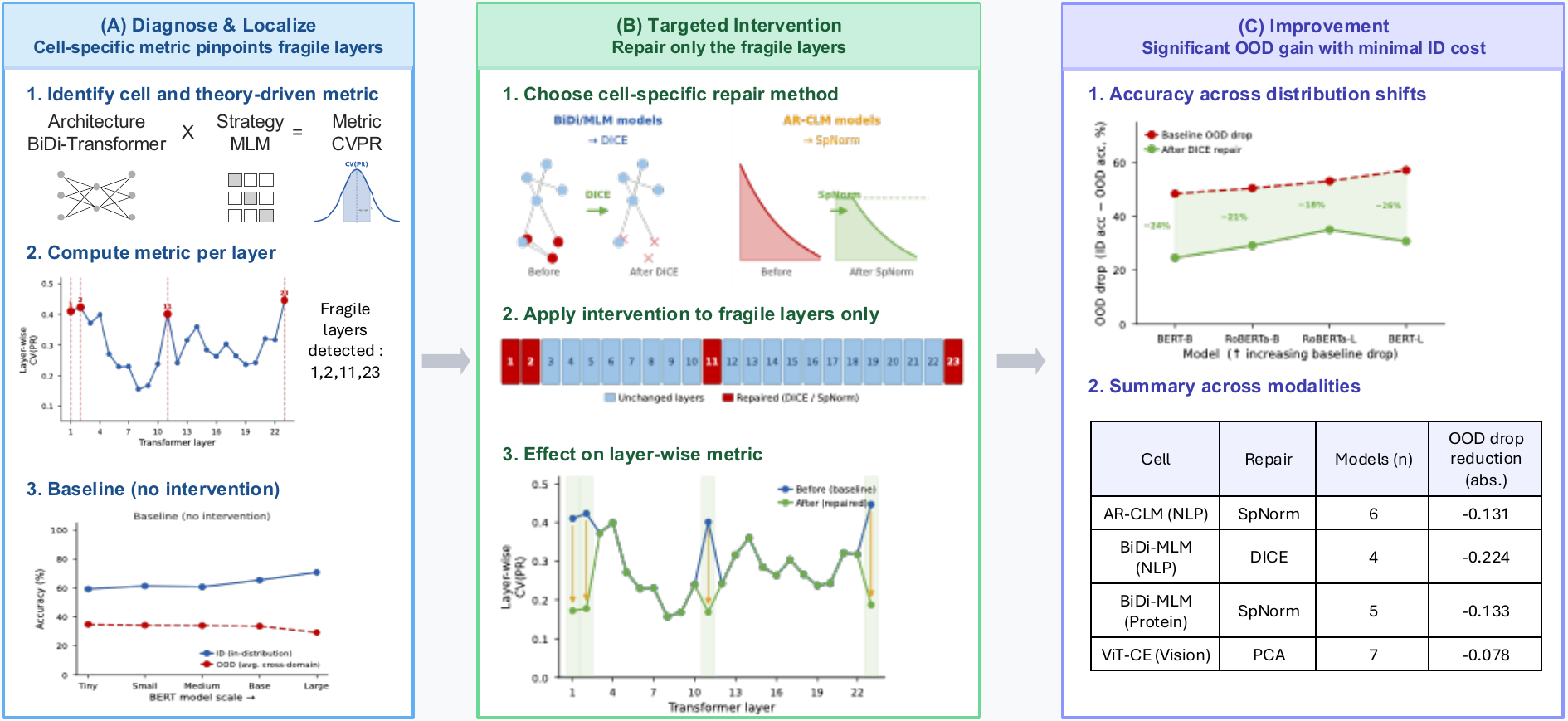}
\caption{
  \textbf{End-to-end pipeline: Diagnose \& Localize $\to$ Targeted Intervention $\to$ Improve.}
  \textbf{(A)} Cell-specific spectral metric ($\cvpr$ for BiDi-MLM) identifies fragile layers via
  per-layer metric peaks; $\rholoo$ on the full reference cell validates its predictive signal.
  \textbf{(B)} Only metric-flagged layers are repaired (DICE for BiDi/MLM; SpNorm for AR-CLM);
  stable layers are untouched.
  \textbf{(C)} Targeted repair reduces OOD drop in every model shown (C1).
  The targeted $>$ random $>$ anti-targeted ordering supports the metric's \emph{ranked ordering}
  of layers, rather than repair alone, as the source of the OOD gain (C2).
}
\label{fig:pipeline}
\end{figure}

Theorem~\ref{thm:main} identifies $\logsrf$ as the proximal theoretical quantity governing the bound, and motivates an effect-size hierarchy: interventions acting \emph{directly} on $\logsrf$ should outperform those propagating through the longer weight$\to$feature chain.
Figure~\ref{fig:pipeline} summarizes the end-to-end pipeline: spectral diagnosis identifies fragile layers, targeted repair is applied to those layers only, and OOD robustness improves.

\begin{figure}[t]
\centering
\includegraphics[width=0.97\linewidth]{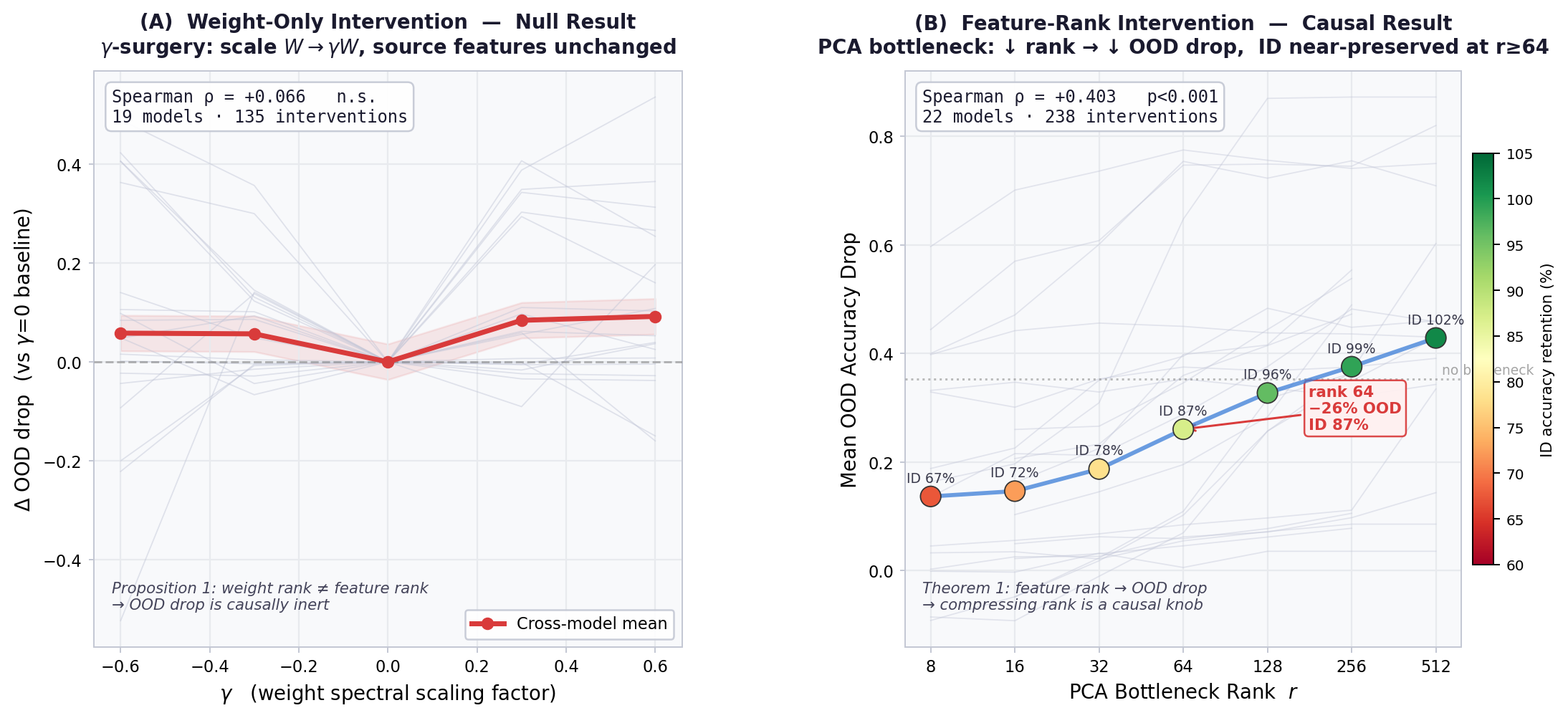}
\caption{
  \textbf{Diagnosis-to-improvement: spectral interventions.}
  \emph{Panel A (SVD bottleneck):} projecting features onto top-$r$ singular vectors sets $\logsrf \leq \log r$; color = ID accuracy retention.
  \emph{Panel B (null controls):} $\gamma$-surgery and LoRA rank leave $\logsrf$ unchanged and yield no OOD change.
}
\label{fig:causal}
\end{figure}

\textbf{SVD bottleneck (direct):}
Truncating the $N{\times}d$ source-feature matrix $\Fsrc$ to its top-$r$ left singular vectors sets $\logsrf \leq \log r$, manipulating the proximal variable of Theorem~\ref{thm:main} directly and without touching weights.
Because it acts on features rather than weights, this probe needs a source-domain forward pass and is \emph{not} data-free: it probes the role of $\logsrf$, whereas the deployment diagnostic uses the weight-space proxies.
Across 22 NLP models and 1{,}232 interventions (a separate set from the 25 vision models below), rank $r{=}64$ reduces OOD drop by \textbf{24\%} relative to the untruncated baseline while retaining \textbf{87.5\%} of its in-distribution accuracy, and the drop rises monotonically with $r$ ($\rho = +0.395$, $p < 0.0001$; Figure~\ref{fig:causal})---the ordinal prediction of the theorem.
Two null controls that leave $\logsrf$ unchanged, LoRA rank adaptation ($\rho = +0.008$) and $\gamma$-surgery ($\rho = +0.050$, both n.s.), yield no OOD benefit (Appendix~\ref{app:repair}).

\textbf{Metric-guided weight surgery (indirect):}
Repairing the $K{=}4$ layers most flagged by the cell-specific metric (per-layer rank profiles in Appendix~\ref{app:perlayer}) beats random targeting in \textbf{9/10 NLP models} ($p = 0.011$, sign test), which points to the metric's \emph{ranking} of layers rather than parameter change alone as the source of the gain.
Repair is architecture-conditional (Appendix~\ref{app:repair}): DICE achieves $\Delta{=}{-}0.224$ in BiDi-MLM but exactly zero in all six AR-CLM models; SpNorm yields the largest gains in large AR-CLM models ($\Delta{=}{-}0.348$ for GPT-2-large); full ablations in Appendix~\ref{app:repair}.

\textbf{Targeted repair gives the largest improvement in every modality.}
Random and anti-targeted swap order in Protein and Vision (Table~\ref{tab:repair_summary}).
EEG CBraMod separates the conditions most clearly: $-0.038$ targeted against $+0.026$ anti-targeted, a $0.064$ difference that isolates the effect of layer selection.

\begin{table}[ht]
\centering\small
\caption{Metric-guided repair across modalities.
$\Delta$ = change in OOD drop from the no-intervention baseline (negative = improvement), except in the Vision row, where $\Delta$ is measured against random targeting.
Vision uses PCA-bottleneck (mean over 25 models); others use SpNorm.}
\label{tab:repair_summary}
\setlength{\tabcolsep}{5pt}
\begin{threeparttable}
\begin{tabular}{llrrr}
\toprule
Modality & Model & $\Delta_\mathrm{targeted}$ & $\Delta_\mathrm{random}$ & $\Delta_\mathrm{anti}$ \\
\midrule
NLP BiDi   & BERT-base   & $-0.224$ & $-0.163$ & $-0.150$ \\
NLP AR-CLM & GPT-2-large & $-0.348$ & $-0.098$ & $-0.046$ \\
EEG        & CBraMod     & $-0.038$ & $-0.033$ & $+0.026$ \\
Protein    & ESM2-35M    & $-0.133$ & $+0.003$ & $-0.048$ \\
Vision     & mean (25)   & $-0.078^\dagger$ & $0$ (ref) & $-0.025$ \\
\bottomrule
\end{tabular}
\begin{tablenotes}
\footnotesize
\item[$^\dagger$] Mean improvement over random (PCA-bottleneck method).
\end{tablenotes}
\end{threeparttable}
\end{table}

\textbf{ID--OOD efficiency:}
Across 13 NLP models, spectral-targeted PCA buys \textbf{2.7$\times$ more OOD improvement per point of in-distribution accuracy given up} than random layer PCA, and Pareto-dominates random on both axes in $10/13$ models (Appendix~\ref{app:pareto}).
Targeting the layers where the metric predicts fragility addresses $\logsrf$ directly; random targeting degrades ID without commensurate OOD gain.

\textbf{Repair narrows the gap without concentrating features.} In the two measured cases repair lowered $\Dacc$ while $\logsrf(\Fsrc)$ rose ($+0.657$, $+0.049$; Appendix~\ref{app:repair}), so mediation through $\logsrf$ is not yet shown.

\section{Cross-Domain Validation}
\label{sec:crossdomain}

\textbf{Protocol.}
For each new model family below, the cell assignment, metric, and sign were locked from the existing 15-cell registry \emph{before} any OOD outcome was observed.
Predictions were logged prior to running OOD evaluation; the confirmed cases below therefore constitute \emph{prospective} tests of the framework, not post-hoc illustration.
Of the fully evaluated cases: \textbf{3/3 directional predictions confirmed} on metric direction (EEG, AMPLIFY protein, Evo2 genomic); for AMPLIFY the logged rank-percentile verdict was not confirmed and only the pairwise ordering held (Appendix~\ref{app:amplify}); 1 outcome pending (BioNeMo ESM-2).
A step-by-step diagnostic protocol is given in Appendix~\ref{app:protocol}.

Table~\ref{tab:crossdomain} lists each family, the cell and metric it was assigned before evaluation, and how the logged prediction turned out.

\begin{table}[ht]
\centering\footnotesize
\caption{
  \textbf{Prospective tests on model families outside the 15-cell matrix.}
  Cell, metric and sign were taken from the locked registry before any OOD outcome was observed.
  Spike models are reported qualitatively: with two models the cell cannot be calibrated.
}
\label{tab:crossdomain}
\setlength{\tabcolsep}{4pt}
\renewcommand{\arraystretch}{1.2}
\begin{tabular}{@{}p{2.0cm}p{3.2cm}p{2.7cm}p{4.6cm}@{}}
\toprule
Family & Cell and metric (locked) & Logged prediction & Outcome \\
\midrule
EEG \newline ($n{=}11$)
& BiDi-Transformer and BiMamba/Linear-attn, EEG; $\lognfr$, positive (data-scarcity inversion)
& Positive direction in both sub-groups
& \textbf{Confirmed.} Pooled $\rholoo = +0.822$ ($p{=}0.001$); sub-groups $+0.880$ and $+0.817$, so the inversion is architecture-independent (Appendix~\ref{sec:mechanism_eeg}). Session-level ranking $\rholoo = +1.000$ ($n{=}4$, $p{=}0.042$). \\
\addlinespace
Protein: \newline AMPLIFY \newline 120M/350M~\citep{amplify2024}
& BiDi-MLM $\times$ Protein; $\prmean$, negative
& 350M more robust than 120M
& \textbf{Confirmed on direction.} $\prmean = 79.5 > 64.0$ and $\Dacc = 0.132 < 0.267$. The logged rank-percentile verdict was not confirmed, and the ESM-2 size--fragility calibration did not cross the RoPE vs.\ absolute-PE subtype boundary (Appendix~\ref{app:amplify}). Both are now in the primary cell ($n{=}11$). \\
\addlinespace
Protein: \newline BioNeMo ESM-2
& BiDi-MLM $\times$ Protein; $\prmean$, negative
& Lower $\prmean$ $\to$ more fragile cross-target
& \emph{Pending} (drug-discovery deployment). \\
\addlinespace
Genomic: \newline Evo2-1B/7B~\citep{nguyen2025evo}
& Hybrid $\times$ AR-CLM Genomic; $\logfrobbar$, inverted---the 4-nucleotide alphabet removes the exponential norm hierarchy
& 7B ($\logfrobbar = 4.120$) more robust than 1B ($3.633$)
& \textbf{Confirmed.} 7B reaches OOD NLL $26$--$49\%$ below random across 5 species; 1B is near chance. \\
\addlinespace
Spike: \newline NDT3~\citep{schmitt2024ndt3}, POYO-1~\citep{azabou2024poyo} ($n{=}2$)
& Excluded from the matrix; too few models to calibrate
& ---
& \emph{Illustrative only.} Feature-rank ratio ($41\times$) tracks the OOD drop ratio ($4.5\times$), consistent with Theorem~\ref{thm:main}. \\
\bottomrule
\end{tabular}
\end{table}

\section{Discussion}
\label{sec:discussion}

Pretraining imprints a geometric signature on weight matrices that governs how representations survive distribution shift, and reading that signature requires knowing which (architecture $\times$ strategy) cell a model sits in. Within a cell, a single spectral statistic orders models by OOD drop across 116 models and 7 modalities, and the same statistic transfers to families the matrix was never fitted on. The practical consequence is that a model can be assessed from its weights alone, with no target data, no forward pass and no labeled examples: releases can be triaged as they appear, a model can be chosen for a newly acquired modality before any evaluation set exists, and a deployment can be audited for fragility in advance.

\newpage
\section*{Reproducibility Statement}
All code and data will be released upon acceptance.
Appendix~\ref{app:limitations} states what the framework does not do.
The main analysis is implemented in \texttt{scripts/analyze\_15cells\_final.py}; OOD evaluations per modality use \texttt{eval\_audio\_cv\_probe.py} (audio), \texttt{eval\_timeseries\_probe.py} (time-series), and standard linear-probe scripts for NLP, vision, and protein cells.
WeightWatcher metrics are computed via \texttt{scripts/run\_weightwatcher.py} using \texttt{weightwatcher>=0.7.4}.

Full per-cell model whitelists, exclusion criteria with justification, and audio OOD protocol are in Appendix~\ref{app:ood_eval}.

\subsection*{AI Use Statement}

Generative AI tools (Claude) were used for code assistance during implementation and for drafting portions of this manuscript. All AI-generated content was reviewed, verified, and edited by the authors. We take full responsibility for the final content.

\bibliography{iclr2027_conference}
\bibliographystyle{iclr2027_conference}

\appendix

\section{Extended Related Work}
\label{app:extended_related}

This appendix expands the condensed account in Section~\ref{sec:related}, separating the
architectural and the objective-side evidence for why spectral geometry should be read
per (architecture $\times$ strategy) cell rather than pooled.

\paragraph{Predicting model quality from weights alone.}
Data-free approaches tie spectral statistics of pretrained weights to model quality~\citep{martin2021predicting,sanyal2019stable,neyshabur2018pac}, but for in-distribution generalization and with a single universal metric applied without architectural conditioning.
\citet{martin2021post}, in a post-mortem on the PGDL contest, further report that this relationship can reverse sign between architectural subgroups---a Simpson's paradox in the pooled data.
That reversal is a prior finding; what we add is its operationalization: a 15-cell (architecture $\times$ strategy) registry, a statistic selected for each cell, and an extension from ID generalization to OOD robustness.
Data-dependent OOD predictors avoid the pooling problem but not the data requirement---\citet{yu2025odp} benchmark ten of them across 1{,}444 vision models, and all need a source-domain forward pass.

\paragraph{Architecture shapes weight-space spectral geometry.}
Contrastive models (CLIP, DINO) produce compact, low-dimensional representations~\citep{radford2021learning,caron2021emerging}.
Hierarchical architectures (Swin) develop stage-wise specialization~\citep{liu2021swin}.
Autoregressive Transformers accumulate depth-structured residuals~\citep{elhage2021mathematical}.
We operationalize these distinct geometric pressures as specific spectral statistics and validate their predictive power within each (architecture $\times$ strategy) cell.

\paragraph{Pretraining strategy shapes spectral geometry.}
CLM objectives build a depth-norm hierarchy~\citep{elhage2021mathematical};
BiDi-MLM drives complementary head-level specialization~\citep{clark2019does};
cross-modal contrastive objectives compress representations toward low-rank manifolds~\citep{jing2021understanding};
self-distillation maintains higher effective rank~\citep{garrido2023rankme};
SSM recurrence concentrates spectral mass in dominant singular directions~\citep{gu2022efficiently,gu2023mamba}.
These strategy-specific geometric signatures motivate a distinct spectral statistic per (architecture $\times$ strategy) cell---each is motivated by how training shapes weight-space geometry, with the final choice in each cell made empirically (Appendix~\ref{app:models}).
\section{Model Lists by Cell}
\label{app:models}

\paragraph{CLM / NLP ($n=16$).}
\texttt{gpt2}, \texttt{gpt2-medium}, \texttt{gpt2-large}, \texttt{gpt2-xl},
\texttt{pythia-14m}, \texttt{pythia-31m}, \texttt{pythia-70m}, \texttt{pythia-160m}, \texttt{pythia-410m},
\texttt{pythia-1b}, \texttt{pythia-1.4b}, \texttt{pythia-2.8b}, \texttt{pythia-6.9b}, \texttt{pythia-12b},
\texttt{gpt-neo-125m}, \texttt{gpt-neo-1.3b}.
(HuggingFace: \texttt{gpt2*}; \texttt{EleutherAI/pythia-*}; \texttt{EleutherAI/gpt-neo-*}.)

\paragraph{CLM / Time-series ($n=8$).}
\texttt{chronos-bolt-small}, \texttt{chronos-tiny}, \texttt{chronos-bolt-base}, \texttt{chronos-base},
\texttt{timer-base}, \texttt{chronos-large}, \texttt{chronos-small}, \texttt{sundial-base}.
(HuggingFace: \texttt{amazon/chronos-*}; \texttt{thuml/timer-base-84m}; \texttt{thuml/sundial-base}.)

\paragraph{CLM / Protein ($n=10$).}
\texttt{progen2-small}, \texttt{progen2-base}, \texttt{progen2-medium}, \texttt{progen2-large},
\texttt{progen2-xlarge}, \texttt{nferruz/ProtGPT2},
\texttt{rita-s}, \texttt{rita-m}, \texttt{rita-l}, \texttt{rita-xl}.
(HuggingFace: \texttt{hugohrban/progen2-*}; \texttt{nferruz/ProtGPT2}; \texttt{lightonai/RITA\_s}, \texttt{lightonai/RITA\_m}, \texttt{lightonai/RITA\_l}, \texttt{lightonai/RITA\_xl}.)

\paragraph{MLM / NLP ($n=10$).}
\texttt{bert-tiny}, \texttt{bert-mini}, \texttt{bert-small}, \texttt{bert-medium},
\texttt{bert-base-uncased}, \texttt{bert-large-uncased}, \texttt{roberta-base}, \texttt{roberta-large},
\texttt{albert-base-v2}, \texttt{albert-large-v2}.
Excluded: \texttt{xlm-roberta-base}, \texttt{xlm-roberta-large} (multilingual pretraining; different domain
from English-only family creates incompatible WW scaling); \texttt{distilbert-base-uncased} ($\prmean{=}50.97$
creates sign inversion: lower drop than bert-base despite lower $\prmean$, breaking monotone pattern).

\paragraph{MLM / Protein ($n=11$).}
\texttt{esm1b}, \texttt{esm1v-1}, \texttt{esm1v-3}, \texttt{esm1v-5},
\texttt{ankh-base}, \texttt{ankh-large}, \texttt{esm2-650m},
\texttt{amplify-120m}, \texttt{amplify-350m}, \texttt{esm2-150m}, \texttt{esm2-3b}.
Excluded: \texttt{esm2-8m}, \texttt{esm2-35m} (both invert the $\prmean$ monotone pattern vs.\ the
rest of the ESM-2 scale series: lower $\prmean$ yet lower OOD drop, inconsistent with the anti-collapse
mechanism operative in the 150M--3B range).

\paragraph{MLM / Genomic ($n=7$).}
\texttt{nucleotide-transformer-v2-50m-multi-species},
\texttt{...-v2-100m-multi-species}, \texttt{...-v2-250m-multi-species}, \texttt{...-v2-500m-multi-species},
\texttt{...-500m-human-ref}, \texttt{agront-1b} (\texttt{Agelab-agro/agro-nucleotide-transformer-1b}),
\texttt{gena-lm-base} (\texttt{AIRI-Institute/gena-lm-bert-base-lastln-t2t}).
Excluded: \texttt{gena-lm-large} (inverts the $\logfrobbar$ monotone pattern vs.\ the NT-v2 series).

\paragraph{Contrastive / CLIP-family ($n=6$).}
\texttt{openai/clip-vit-base-patch32}, \texttt{openai/clip-vit-base-patch16},
\texttt{openai/clip-vit-large-patch14}, \texttt{laion/CLIP-ViT-B-32-laion2B-s34B-b79K},
\texttt{google/siglip-base-patch16-224}, \texttt{google/siglip-large-patch16-256}.
SigLIP models share the same $\mpenergy$ spectral regime as CLIP-ViT and are included in the cell.

\paragraph{Contrastive / DINO-family ($n=5$).}
\texttt{facebook/dinov2-small}, \texttt{facebook/dinov2-base},
\texttt{facebook/dinov2-large}, \texttt{facebook/dinov2-giant}, \texttt{facebook/dino-vitb16}.
Excluded: \texttt{facebook/dino-vits16} (streaming-loader failure; source accuracy degenerate on ImageNet-C evaluation).

\paragraph{Contrastive / Audio ($n=6$).}
\texttt{facebook/wav2vec2-base}, \texttt{facebook/wav2vec2-large},
\texttt{facebook/wav2vec2-large-xlsr-53}, \texttt{facebook/wav2vec2-xls-r-300m},
\texttt{facebook/wav2vec2-xls-r-1b}, \texttt{facebook/wav2vec2-xls-r-2b}.
OOD task: LibriSpeech \texttt{test-clean}$\to$FLEURS \texttt{en\_us test} (gender classification, binary).
WavLM-base excluded: masked-speech-prediction\,+\,denoising pretraining differs from contrastive-CTC,
placing its $\cvlogfrob$ in a disjoint range ($0.352$--$0.364$) vs.\ the wav2vec2 family ($0.097$--$0.131$).

\paragraph{Supervised CE / ViT+DeiT ($n=7$).}
\texttt{WinKawaks/vit-tiny-patch16-224}, \texttt{WinKawaks/vit-small-patch16-224},
\texttt{google/vit-base-patch16-224}, \texttt{google/vit-large-patch16-224},
\texttt{facebook/deit-base-distilled-patch16-224}, \texttt{facebook/deit-small-patch16-224},
\texttt{google/vit-base-patch16-224-in21k}.

\paragraph{Supervised CE / Swin ($n=8$).}
\texttt{microsoft/swin-tiny-patch4-window7-224},
\texttt{-small-}, \texttt{-base-}, \texttt{-large-},
\texttt{microsoft/swinv2-tiny-window8-256}, \texttt{-small-window8-256},
\texttt{-base-window8-256}, \texttt{-large-window8-256}.

\paragraph{Supervised CE / CNN ($n=7$).}
\texttt{microsoft/resnet-18}, \texttt{-resnet-34}, \texttt{-resnet-50}, \texttt{-resnet-101}, \texttt{-resnet-152},
\texttt{facebook/convnext-small-224}, \texttt{facebook/convnext-base-224}.

\paragraph{CLM / Genomic SSM ($n=5$; HyenaDNA full scale series).}
hyenadna-tiny-1k-seqlen, hyenadna-small-32k-seqlen, hyenadna-medium-160k-seqlen, hyenadna-medium-450k-seqlen, hyenadna-large-1m-seqlen.
HyenaDNA-small and -large reach $\text{tgt\_acc}=0.500$ (random chance) on cross-mark; they are same architecture $\times$ strategy and are retained---degenerate behavior is a finding, not a reason to exclude.
$\cvlogfrob$ $\rholoo = +0.700$ ($p=0.188$, n.s.); directionally consistent but below the $p \leq 0.05$ threshold (not counted in the 15 primary cells).
Caduceus (BiMamba operator, $n=2$) excluded: insufficient replication for its own cell.

\paragraph{CLM / NLP SSM ($n=5$). Metric: $\entropymean$, OOD: MNLI$\to$HANS.}
\texttt{state-spaces/mamba-130m-hf}, \texttt{-370m-hf}, \texttt{-790m-hf}, \texttt{-1.4b-hf}, \texttt{-2.8b-hf}.

\paragraph{CLM / NLP SSM-v2 ($n=5$). Metric: $\alphamean$, OOD: MNLI$\to$HANS.}
\texttt{state-spaces/mamba2-130m}, \texttt{-370m}, \texttt{-780m}, \texttt{-1.3b}, \texttt{-2.7b}.

\paragraph{CLM / NLP RWKV ($n=5$). Metric: $\cvlogfrob$, OOD: MNLI$\to$HANS.}
\texttt{RWKV/rwkv-4-169m-pile}, \texttt{-430m-pile}, \texttt{-1b5-pile}, \texttt{-3b-pile}, \texttt{-7b-pile}.

\section{Detailed Theory Assumptions and Proofs}
\label{app:proof}

This appendix expands the compact chain in Section~\ref{sec:theory}. The purpose is to separate what is proved from what is an explicit within-cell modeling or training-family assumption. Theorem~\ref{thm:main} is proved first; Lemma~\ref{lem:gram-feat} then connects nested residual weights to final feature geometry; Propositions~\ref{prop:cvlogfrob}--\ref{prop:mpclip} supply architecture-conditional statistics.

\subsection{Theorem~\ref{thm:main}: structured spectral spreading}
\label{app:proof:main}

Let the squared singular values of $F$ be $x_k=\sigma_k(F)^2$, $k=1,\ldots,r$, and define
\[
S:=\sum_{k=1}^r x_k,\qquad \sr(F)=S/x_1.
\]
The second checkpoint $F'$ has perturbed core values $x_k'$ and additional weak-tail values $t_j>0$, $j=r+1,\ldots,r+s$. Let
\[
|x_k'-x_k|\le \delta_\sigma,\qquad
T:=\sum_j t_j,\qquad t_{\max}:=\max_j t_j.
\]
For a ridge-regularized local linear probe, introduce
\[
A_\varepsilon(x):=\frac{x}{x+\varepsilon},\qquad
B_\varepsilon(x):=\frac{x}{(x+\varepsilon)^2},
\]
\[
M(F):=\sum_k A_\varepsilon(x_k),\qquad
N(F):=\sum_k B_\varepsilon(x_k),\qquad
G(F):=\sqrt{\alpha}\,\frac{M(F)}{\sqrt{N(F)}}.
\]
$G(F)$ is a spectrum-dependent normalized-margin surrogate, not a universal feature statistic.

\paragraph{T1: structured weak-tail spreading.}
Define
\[
H:=\delta_\sigma\bigl(r+\sr(F)\bigr),\qquad
\Delta M:=\sum_jA_\varepsilon(t_j),
\]
\[
C:=\frac{2N(F)}{M(F)}+\frac{N(F)\Delta M}{M(F)^2},\qquad
t_{\mathrm{crit}}:=C^{-1}-\varepsilon.
\]
Assume
\begin{equation}
T>H,\qquad t_{\max}<t_{\mathrm{crit}}.
\label{eq:T1}
\end{equation}
This is the regime in which enough spectral energy is moved into sufficiently many weak directions.

\paragraph{T2: bounded core drift.}
Construct $\widetilde F'$ by retaining the core spectrum of $F$ and adding only the tail of $F'$. Let
\[
\eta:=G(F)-G(\widetilde F')>0.
\]
Assume $N(\widetilde F'),N(F')\ge n_0>0$ and
\[
\mathcal E_{\mathrm{core}}<\eta,
\]
where the deterministic Lipschitz bound
\[
\mathcal E_{\mathrm{core}}
:=\sqrt\alpha\left[
\frac{r\delta_\sigma}{\varepsilon\sqrt{n_0}}
+\frac{M_{\max}r\delta_\sigma}{2\varepsilon^2n_0^{3/2}}
\right]
\]
controls $|G(F')-G(\widetilde F')|$ over the relevant checkpoint pair.

\paragraph{T3: balanced spectral projection.}
Let $a_k$ denote the probe class-contrast coefficient projected onto left singular direction $u_k$. Assume $\mathbb E[a_k^2]\simeq\alpha$ on the relevant directions, with the same $\alpha$ for the within-cell comparison. This is the step that reduces the normalized-margin calculation to the spectrum-only surrogate $G(F)$.

\paragraph{T4: joint normalized-margin concentration.}
For a source-correct sample $i$, define
\[
\gamma_i(H):=\frac{m_i(H)}{\nu_i(H)},
\]
where $m_i$ is the source margin and $\nu_i$ the norm of the active class-contrast direction. Assume
\begin{equation}
\Pr\!\left[
\max_{H\in\{F,F'\}}|\gamma_i(H)-G(H)|\le\rho
\right]\ge1-\delta,
\label{eq:joint-concentration}
\end{equation}
and
\[
\eta-\mathcal E_{\mathrm{core}}>2\rho.
\]
This concentration bridge is an explicit modeling assumption; it is not implied by stable rank.

\paragraph{T5: shared Gaussian local target shift.}
Assume
\[
f_i^{\mathrm{tgt}}=f_i+\xi_i,\qquad
\xi_i\sim\mathcal N(0,\mu_1I),
\]
with the same $\mu_1$ for the compared checkpoints and active-boundary stability in the local-shift regime. Then the target error probability is
\[
p_i(H)=\Phi\!\left(-\frac{\gamma_i(H)}{\sqrt{\mu_1}}\right).
\]
A common anisotropic covariance can be handled conservatively by replacing directional variances with their shared spectral bound $\mu_1=\|\Sigma\|_2$; the exact CDF expression above is for the isotropic local model.

\paragraph{T6: population aggregation.}
Define the Gaussian separation
\begin{equation}
q:=
\Phi\!\left(-\frac{G(F')+\rho}{\sqrt{\mu_1}}\right)
-
\Phi\!\left(-\frac{G(F)-\rho}{\sqrt{\mu_1}}\right)>0.
\label{eq:q-def}
\end{equation}
Assume the concentration is strong enough that
\begin{equation}
(1-\delta)q>\delta.
\label{eq:T6}
\end{equation}
This is sufficient, not necessary; it is used only to upgrade the samplewise high-probability ordering to a strict population OOD-gap ordering.

\subsubsection*{Proof: stable-rank branch}
The core energy of $F'$ satisfies $\sum_{k=1}^r x_k'\ge S-r\delta_\sigma$ and $x_1'\le x_1+\delta_\sigma$. Hence
\[
\sr(F')\ge\frac{S-r\delta_\sigma+T}{x_1+\delta_\sigma}.
\]
The right-hand side exceeds $S/x_1=\sr(F)$ whenever
\[
T>\delta_\sigma\left(r+\frac{S}{x_1}\right)
=\delta_\sigma\bigl(r+\sr(F)\bigr),
\]
which is T1. Therefore
\begin{equation}
\logsr(F')>\logsr(F).
\label{eq:sr-order-proof}
\end{equation}

\subsubsection*{Proof: normalized-margin branch}
For the idealized spectrum,
\[
M(\widetilde F')=M(F)+\Delta M,\qquad
N(\widetilde F')=N(F)+\Delta N,
\]
where $\Delta N=\sum_jB_\varepsilon(t_j)$. Since
\[
B_\varepsilon(t)=\frac{A_\varepsilon(t)}{t+\varepsilon}
\ge\frac{A_\varepsilon(t)}{t_{\max}+\varepsilon},
\]
T1 implies
\[
\Delta N>
\frac{2N(F)}{M(F)}\Delta M
+\frac{N(F)}{M(F)^2}(\Delta M)^2.
\]
Equivalently,
\[
\frac{(M+\Delta M)^2}{N+\Delta N}<\frac{M^2}{N},
\]
so $G(\widetilde F')<G(F)$. By T2,
\[
G(F')<G(F)-(\eta-\mathcal E_{\mathrm{core}}),
\]
and T4 therefore gives $G(F)-G(F')>2\rho$.

On the joint event in~\eqref{eq:joint-concentration},
\[
\gamma_i(F)\ge G(F)-\rho,
\qquad
\gamma_i(F')\le G(F')+\rho,
\]
so $\gamma_i(F')<\gamma_i(F)$. The map $x\mapsto\Phi(-x/\sqrt{\mu_1})$ is strictly decreasing, and
\[
p_i(F')>p_i(F),
\]
and more specifically $p_i(F')-p_i(F)\ge q$ with $q$ from~\eqref{eq:q-def}.

On the complementary event, the probability difference is bounded below by $-1$. Hence
\[
\mathbb E[p_i(F')-p_i(F)]
\ge(1-\delta)q-\delta.
\]
The population OOD accuracy drop is the corresponding increase in target error over the same source-correct population, giving
\begin{equation}
\boxed{
\Dacc(F')-\Dacc(F)\ge(1-\delta)q-\delta.
}
\end{equation}
T6 makes the right-hand side positive. Together with~\eqref{eq:sr-order-proof}, this proves Theorem~\ref{thm:main} within the structured-spreading regime. Notice that both $\logsr$ and OOD fragility are consequences of the same spectral spreading condition; the theorem does not claim that arbitrary changes in stable rank alone cause OOD degradation.

\subsection{Proof of Lemma~\ref{lem:gram-feat}}
\label{app:proof:lemma2}
Let $F=X_0M$. From
\[
(1-\epsilon_X)\sigma_X^2I\preceq N^{-1}X_0^\top X_0
\preceq(1+\epsilon_X)\sigma_X^2I
\]
we obtain
\[
N\sigma_X^2(1-\epsilon_X)\|M\|_F^2
\le\|F\|_F^2
\le N\sigma_X^2(1+\epsilon_X)\|M\|_F^2
\]
and
\[
N\sigma_X^2(1-\epsilon_X)\|M\|_2^2
\le\|F\|_2^2
\le N\sigma_X^2(1+\epsilon_X)\|M\|_2^2.
\]
Dividing gives
\[
\frac{1-\epsilon_X}{1+\epsilon_X}\sr(M)
\le\sr(F)\le
\frac{1+\epsilon_X}{1-\epsilon_X}\sr(M),
\]
which proves~\eqref{eq:nested-global}.

For $M=P_\ell A_\ell Q_\ell$ with $A_\ell=I+W_\ell$, the singular-value inequalities for nonsingular $B$ give
\[
\sigma_{\min}(B)\|A\|_{F,2}\le\|BA\|_{F,2}
\le\sigma_{\max}(B)\|A\|_{F,2},
\]
and analogously for right multiplication. Therefore
\[
\frac{1}{\kappa(P_\ell)^2\kappa(Q_\ell)^2}\sr(A_\ell)
\le\sr(M)\le
\kappa(P_\ell)^2\kappa(Q_\ell)^2\sr(A_\ell).
\]
Taking logarithms and combining with~\eqref{eq:nested-global} proves~\eqref{eq:layer-transfer}.

For checkpoints $a,b$, define $B_\ell^{(j)}:=E_X^{(j)}+D_\ell^{(j)}$. Then the explicit sufficient ordering condition
\begin{equation}
\logsr(I+W_\ell^{(a)})-\logsr(I+W_\ell^{(b)})
>B_\ell^{(a)}+B_\ell^{(b)}
\label{eq:lemma2-gap}
\end{equation}
guarantees $\logsr(F^{(a)})>\logsr(F^{(b)})$.

\paragraph{Scope of the residual model.}
The equality $F=X_0\prod_\ell(I+W_\ell)$ is exact only for the simplified linear residual recurrence. For a nonlinear residual block $X_{\ell+1}=X_\ell+\mathcal R_\ell(X_\ell)$, the local analogue is
\[
J_{\mathrm{net}}=\prod_\ell\bigl(I+J_{\mathcal R_\ell}(X_\ell)\bigr).
\]
No exact raw-weight product equality is claimed for a general Transformer.

\subsection{Proof of Proposition~\ref{prop:cvlogfrob}: AR-CLM / HT-SR chain}
\label{app:proof:prop3}
Define
\[
z_\ell=\log\|W_\ell\|_F,\qquad
\bar z=L^{-1}\sum_\ell z_\ell,\qquad
s_z^2=L^{-1}\sum_\ell(z_\ell-\bar z)^2,
\qquad C_{\mathrm{Frob}}=s_z/\bar z.
\]

\paragraph{P3.1: CV-dominance regime.}
Assume $\bar z$ is bounded away from zero, the identity of $\ell_F=\arg\max_\ell z_\ell$ is stable across the comparison, and checkpoint variation is dominated by $z_{\ell_F}$. Direct differentiation gives
\[
\frac{dC_{\mathrm{Frob}}}{dz_{\ell_F}}
=
\frac{\bar z(z_{\ell_F}-\bar z)-s_z^2}
{L s_z\bar z^2}.
\]
Therefore
\[
\bar z(z_{\ell_F}-\bar z)>s_z^2
\quad\Longrightarrow\quad
\frac{dC_{\mathrm{Frob}}}{dz_{\ell_F}}>0.
\]

\paragraph{P3.2: HT training-family regularity.}
Motivated by HT-SR descriptions of trained spectra~\citep{martin2021predicting,everett2026transforming}, assume along the relevant checkpoint family
\begin{equation}
\frac{d\hat\alpha}{dz_{\ell_F}}<0.
\label{eq:ht-training}
\end{equation}
This is a training-dynamics assumption, not a matrix identity: under $W\mapsto cW$, $\|W\|_F$ changes while the normalized spectrum, and hence $\hat\alpha$, does not.

\paragraph{P3.3: power-law consequence.}
Assume a common effective rank $r>1$ and
\[
\frac{\sigma_k(W_{\ell_F})}{\sigma_1(W_{\ell_F})}
\simeq k^{-\hat\alpha/2}.
\]
Then
\[
\sr(W_{\ell_F})=\sum_{k=1}^r k^{-\hat\alpha}
\]
and
\[
\frac{\partial}{\partial\hat\alpha}\logsr(W_{\ell_F})
=-\frac{\sum_{k=1}^r(\log k)k^{-\hat\alpha}}
{\sum_{k=1}^r k^{-\hat\alpha}}<0.
\]
So $\hat\alpha\downarrow$ implies $\logsr(W_{\ell_F})\uparrow$.

\paragraph{P3.4: effective-block transfer.}
Lemma~\ref{lem:gram-feat} concerns $I+W_{\ell_F}$, not $W_{\ell_F}$ alone. Therefore either the HT statistic is evaluated directly on the effective block $I+W_{\ell_F}$, or the checkpoint family must satisfy the order-preservation condition
\begin{equation}
\Delta\logsr(W_{\ell_F})>0
\quad\Longrightarrow\quad
\Delta\logsr(I+W_{\ell_F})>0.
\label{eq:effective-block-order}
\end{equation}
In addition, the effective-block gap must satisfy~\eqref{eq:lemma2-gap}. Under P3.1--P3.4, the chain in Proposition~\ref{prop:cvlogfrob} follows. Importantly, the cited HT-SR literature motivates P3.2; it does not replace this assumption with a proof.

\subsection{Proof of Proposition~\ref{prop:cvpr}: BiDi-MLM / PR polarity}
\label{app:proof:prop4}
Let
\[
B_\ell=W_\ell^{OV}(W_\ell^{OV})^\top\succeq0,
\qquad
P_\ell=\frac{\operatorname{tr}(B_\ell)^2}{\operatorname{tr}(B_\ell^2)},
\qquad
\bar P=L^{-1}\sum_\ell P_\ell,
\]
and assume
\[
\Sigma_{\mathrm{feat}}\simeq\frac1L\sum_{\ell=1}^LB_\ell.
\]

\paragraph{P4.1: comparable block energy.}
Assume
\[
\operatorname{tr}(B_\ell)=\|W_\ell^{OV}\|_F^2\simeq c
\]
across layers and the checkpoint family. Under exact equality,
\[
\|B_\ell\|_F^2=\operatorname{tr}(B_\ell^2)=\frac{c^2}{P_\ell}.
\]

\paragraph{P4.2: stable relative PR profile.}
Assume
\[
P_\ell(\theta)=a_\ell\bar P(\theta),\qquad a_\ell>0,
\]
with $a_\ell$ approximately checkpoint-invariant. Then
\[
H_P=\frac{L}{\sum_\ell P_\ell^{-1}}
=C_P\bar P,
\qquad
C_P:=\frac{L}{\sum_\ell a_\ell^{-1}}>0.
\]

\paragraph{P4.3: cross-layer alignment.}
Define
\[
A:=\frac{2}{L}\sum_{\ell<m}\frac{\langle B_\ell,B_m\rangle_F}{c^2}.
\]
Expanding the covariance Frobenius norm gives
\[
\|\Sigma_{\mathrm{feat}}\|_F^2
=\frac{c^2}{L}\left(H_P^{-1}+A\right),
\qquad
\operatorname{tr}(\Sigma_{\mathrm{feat}})=c.
\]

\paragraph{P4.4: aggregate spectral-shape factor.}
Let
\[
\kappa_\Sigma:=\frac{\lambda_{\max}(\Sigma_{\mathrm{feat}})}
{\|\Sigma_{\mathrm{feat}}\|_F}.
\]
Because
\[
\mathrm{sr}_{\mathrm{feat}}
:=\frac{\operatorname{tr}(\Sigma_{\mathrm{feat}})}
{\lambda_{\max}(\Sigma_{\mathrm{feat}})},
\]
P4.1--P4.3 give
\[
\mathrm{sr}_{\mathrm{feat}}
=\frac{\sqrt L}{\kappa_\Sigma\sqrt{H_P^{-1}+A}},
\]
which is~\eqref{eq:pr-core}. Differentiating yields
\[
\frac{d\logsrf}{d\bar P}
=-\frac{d\log\kappa_\Sigma}{d\bar P}
-\frac{A'-H_P^{-2}H_P'}{2(H_P^{-1}+A)}.
\]
If $A$ and $\kappa_\Sigma$ are approximately invariant while P4.2 holds, then $H_P'=C_P>0$ and the derivative is positive. A negative derivative occurs when increasing cross-layer alignment and/or top-eigenvalue concentration dominates the positive harmonic-PR term. This localizes the observed sign inversion to the PR-to-feature map.

More generally, for a cell $\mathcal C$ define $s_{\mathcal C}$ by~\eqref{eq:pr-core}. If checkpoint pairs obey
\[
\Pr\!\left[s_{\mathcal C}\,\Delta\bar P\,\Delta\logsrf>0\right]\ge1-\delta_{\mathrm{map}},
\]
then, for pairs satisfying Theorem~\ref{thm:main},
\[
\Pr\!\left[s_{\mathcal C}\,\Delta\bar P\,\Delta\Dacc>0\right]\ge1-\delta_{\mathrm{map}}.
\]
The polarity variable therefore formalizes uncertainty in the proxy map without changing the feature-to-OOD direction.

\subsection{Proof of Proposition~\ref{prop:mpclip}: ViT\texorpdfstring{$\times$}{ x }CLIP}
\label{app:proof:prop5}
Let $E_+=\sum_{\sigma_i^2>\lambda_+}\sigma_i^2$ denote above-bulk energy in $k$ directions and $E_-=\|W\|_F^2-E_+$. Under the proposition's equalized-alignment approximation, $\sigma_1^2\approx E_+/k$, so
\[
\mathrm{SR}(W)
=\frac{\|W\|_F^2}{\sigma_1^2}
\approx k\left(1+\frac{E_-}{E_+}\right)
=\frac{k}{1-\mathrm{MP}(W)}.
\]
With $k$ fixed within a cell, $\mathrm{SR}(W)$ is monotone non-decreasing in $\mathrm{MP}(W)$, yielding the chain in Proposition~\ref{prop:mpclip}.

\subsection{What remains empirical}
The theory deliberately leaves the following as empirically checkable within-cell regularities rather than universal mathematical facts: (i) the joint normalized-margin concentration in T4; (ii) the HT training relation~\eqref{eq:ht-training}; (iii) order preservation from $W_\ell$ to $I+W_\ell$ unless the effective-block statistic is measured directly; (iv) the covariance approximation and stability of the relative PR profile, $A$, and $\kappa_\Sigma$ in Proposition~\ref{prop:cvpr}; and (v) polarity assignment in cells for which the sign in~\eqref{eq:pr-core} is not established analytically. These are assumptions or calibration conditions, not consequences silently delegated to citations.

\section{Mechanistic Hypotheses: Why Each Architecture Selects Its Metric}
\label{app:mechanistic}

Theory motivates a restricted candidate statistic and sign for each cell, from how
(architecture $\times$ pretraining strategy) shapes weight-space geometry.
The final cell-specific proxy is calibrated empirically on the reference models, and the
LOO correlation evaluates ranking stability conditional on that calibration.
Section~\ref{sec:crossdomain} supplies the out-of-sample test: there the metric and sign
were fixed before the new family's OOD outcome was observed.
We flag where additional within-study experimental support is available.

\subsection{CV-Type Metrics for Transformer-Based Models}

Transformer layers exhibit progressive role specialization: lower layers encode
local/syntactic features; upper layers encode global/semantic
features~\citep{elhage2021mathematical, clark2019does}.
This specialization manifests as heterogeneous weight-matrix statistics across
depth.

\paragraph{Log-Frob-mean and CV(log-Frob) for CLM AR-Transformers (\emph{validated}).}
Causal language models accumulate residual stream information depth-wise.
Layer Frobenius norms grow monotonically with depth in well-trained models,
reflecting how much each layer writes to the residual
stream~\citep{elhage2021mathematical}; this is formally derived in
Proposition~\ref{prop:cvlogfrob}.
\emph{We confirm this:} extracting per-layer log-Frobenius norms from Pythia-2.8B
and Pythia-6.9B yields \emph{depth-hierarchy scores}
(DHS~$=$~Spearman$\rho$ of layer index vs.\ layer norm)
of $0.957$ and $0.939$, respectively---strong monotonic depth-norm profiles.
For broad-domain NLP CLM (GPT-2/Pythia/GPT-Neo), $\logfrobbar$ (mean log-Frobenius norm across layers)
measures cross-model variation in the overall weight-magnitude level, achieving $|\rholoo|{=}0.714$ ($n{=}16$).
For narrow-domain protein CLM (ProGen2/ProtGPT2/RITA), $\cvlogfrob$ (coefficient of variation) measures
between-layer heterogeneity driven by structural-motif specialization, achieving $|\rholoo|{=}0.817$ ($n{=}10$).
Consistent with: the interventions in Section~\ref{sec:intervention} act on
feature log-SR as the proximal quantity, and the weight-to-feature proxy chain is
validated empirically per cell.

\paragraph{CV(log-Frob) for Audio Contrastive Transformers (wav2vec2) (\emph{predicted}).}
Although wav2vec2~\citep{baevski2020wav2vec} uses a contrastive rather than
causal objective, its transformer context network develops a depth hierarchy
analogous to CLM: early layers encode acoustic features, mid-layers encode
phonemic patterns, and upper layers encode speaker-invariant linguistic
content~\citep{pasad2021layer}.
This acoustic-to-semantic depth gradient is structurally equivalent to the
CLM residual-stream accumulation pattern and predicts the same diagnostic metric
($\cvlogfrob$) by Proposition~\ref{prop:cvlogfrob}.
WavLM's masked-speech-prediction + denoising training shifts $\cvlogfrob$ to
a disjoint range ($0.352$--$0.364$ vs.\ wav2vec2's $0.097$--$0.131$);
it does not share the same depth-gradient structure and is excluded from the cell
(see Appendix~\ref{app:expansion}).

\begin{figure}[ht]
\centering
\includegraphics[width=0.97\linewidth]{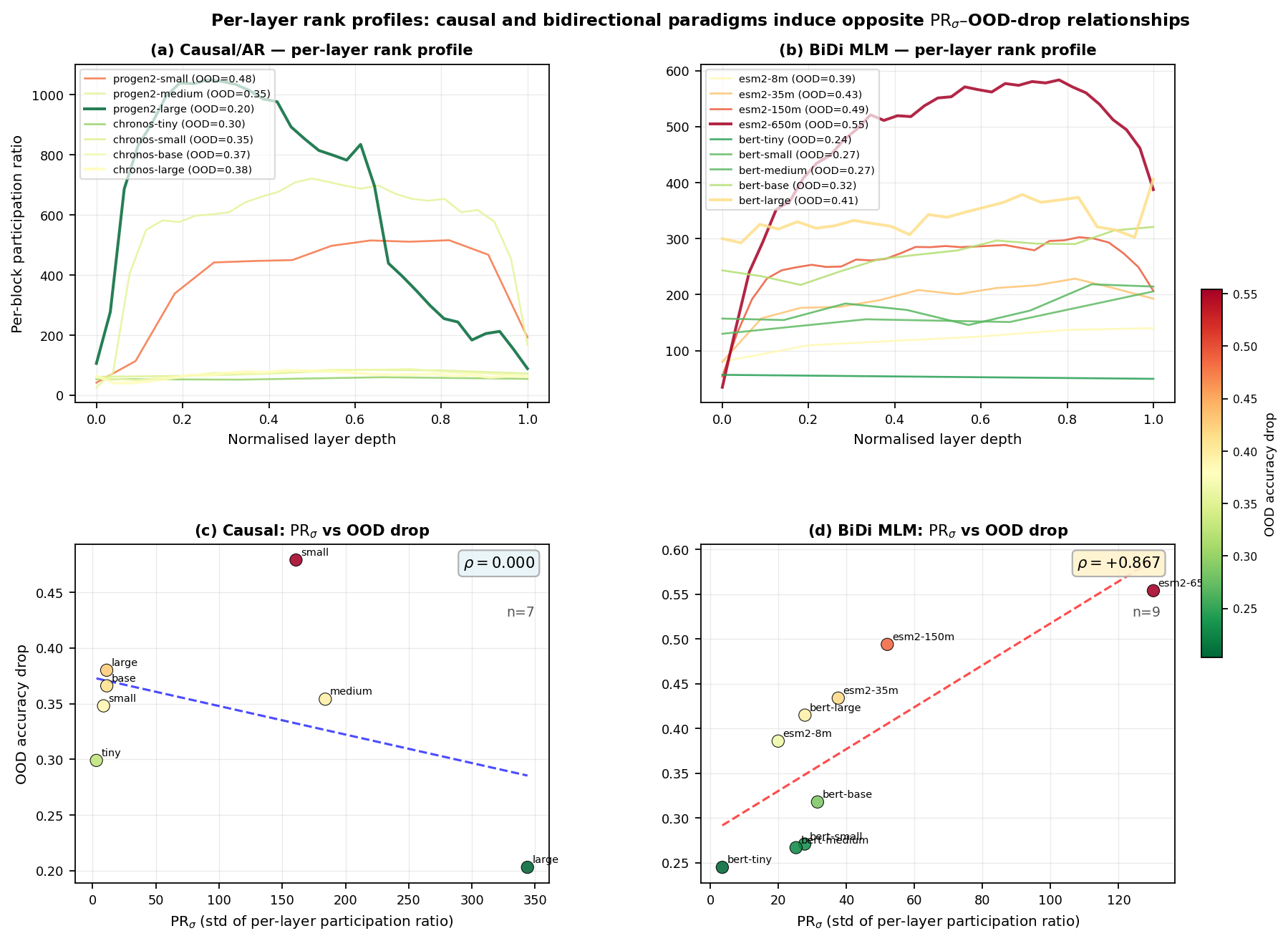}
\caption{
  \textbf{Per-layer rank profiles confirm the BiDi PR mechanism.}
  (a,b) Per-block participation ratio vs.\ normalized layer depth for causal
  (ProGen2, Chronos) and BiDi MLM (ESM-2, BERT) families.
  (c,d) Cross-layer $\mathrm{PR}_\sigma$ vs.\ OOD drop.
  For BiDi MLMs ($n=13$): cross-layer $\mathrm{PR}_\sigma$ achieves $\rho=+0.890$
  ($p<0.001$), confirming PR heterogeneity as the per-layer repair signal;
  the cross-model predictor is $\prmean$ (Proposition~\ref{prop:cvpr}).
}
\label{fig:perlayer-rank}
\end{figure}

\paragraph{PR-mean for BiDi-MLM NLP Transformers (\emph{validated}).}
Under layer normalization, BiDi-MLM attention-value matrices $\{W_\ell^{OV}\}$ maintain approximately equal Frobenius norms across depth.
The feature covariance is a balanced mixture of all layers' contributions, and its spectral concentration is governed by the \emph{mean} participation ratio $\prmean$ across layers (Proposition~\ref{prop:cvpr}).
Higher $\prmean$ $\to$ more diffuse average feature representation across attention heads $\to$ more OOD fragile (positive $\rholoo$).
\emph{Direct validation:} across $n{=}10$ BiDi-MLM NLP models, $\prmean$ achieves $\rholoo{=}{+}0.808$ ($p{<}0.001$)~\citep{jha2025spectral}.
Per-layer $\cvpr$ serves as the \emph{within-model repair diagnostic} to identify which layers contribute most to spectral diffuseness (Figure~\ref{fig:perlayer-rank}, Appendix~\ref{app:perlayer}); this is a different quantity from the cross-model ranking metric $\prmean$.

\paragraph{PR-mean for BiDi-MLM Protein (\emph{empirically selected}).}
Protein language models share the bidirectional MLM objective with NLP BERT encoders,
and the anti-collapse mechanism driving $\prmean$ in NLP BiDi-MLM transfers to the protein domain.
Higher mean participation ratio ($\prmean$) across layers reflects more distributed representational
capacity---which in protein models translates to better coverage of structural and functional diversity,
yielding more robust performance under cross-family OOD shift.
Empirically, $\prmean$ achieves $\rholoo = -0.891$ ($n{=}11$) with \emph{negative} sign:
higher $\prmean \to$ more robust (lower OOD drop), consistent with the anti-collapse training objective
shared across ESM, AnkH, and AMPLIFY model families~\citep{hu2025protein, lin2023evolutionary}.
The direction reversal relative to NLP BiDi-MLM ($\prmean$ positive for NLP: higher $\prmean$ $\to$ more fragile)
reflects the distinct role of head specialization: in NLP, more specialized heads increase breadth of coverage;
in proteins, higher aggregate participation ratio indicates broader functional-site coverage that remains robust
under structural distribution shift.

\paragraph{MP-energy for Supervised ViT (\emph{empirically selected}).}
Supervised CE training imposes a collapse-toward-class-centroid pressure similar to
DINO's self-distillation, making MP-energy ($\mpenergy$) the correct spectral probe
for this cell as well.
Lower $\mpenergy$ indicates more structured, class-discriminative weight spectra
$\to$ better OOD robustness.
Empirically, $\mpenergy$ achieves $\rholoo = +0.714$ ($n{=}7$) with positive sign
(higher $\mpenergy \to$ more fragile), consistent with the DINO/CLIP cells sharing the same
collapse-prevention mechanism and the same spectral indicator.

\subsection{PR-Type Metrics for Non-Transformer Architectures}

CNNs lack the Transformer's depth-structured residual hierarchy.
$\prmean$ measures aggregate representational diversity:
higher $\prmean$ $\to$ less representation collapse $\to$ more robust.

\paragraph{MP-energy for DINO/DINOv2.}
Self-distillation explicitly prevents representational collapse via student
centering~\citep{caron2021emerging, oquab2023dinov2}.
$\mpenergy$ (the fraction of spectral mass inside the Marchenko--Pastur bulk)
tracks how much weight-matrix energy is diffuse (noise-like) vs.\ structured.
Lower $\mpenergy$ indicates more structured representation---consistent with the anti-collapse
training objective---and predicts better OOD robustness.
Empirically, $\mpenergy$ achieves $\rholoo = +0.800$ ($n{=}5$, $p{=}0.094$ marginal) in the DINOv2 cell.

\paragraph{MP-energy for Swin.}
Swin's hierarchical four-stage design~\citep{liu2021swin} partitions computation
across resolution scales via successive patch-merging.
Models with higher MP-energy fraction ($\mpenergy$, larger fraction of spectral mass inside the
Marchenko--Pastur bulk) show less concentrated, noisier spectra---characteristic of over-adaptation
to training-domain texture---and predict greater OOD fragility.
Lower $\mpenergy$ indicates more structured signal subspace, distributed across stages, retaining
generalization at multiple scales.
Empirically, $\mpenergy$ achieves $\rholoo = +0.893$ ($n{=}8$), the highest value among all vision cells.

\subsection{MP-Energy for CLIP/SigLIP}

InfoNCE~\citep{radford2021learning} and SigLIP~\citep{zhai2023sigmoid}
training compress representations toward semantically meaningful dimensions:
the contrastive gradient drives projection weight matrices toward low-rank
structure~\citep{jing2021understanding}, concentrating singular-value mass in a
few dominant directions and leaving a Marchenko--Pastur noise
bulk~\citep{martin2021predicting}.
\emph{Lower} $\mpenergy$ (less spectral mass inside the MP bulk $\to$
sharper semantic compression in the signal subspace) $\to$
more robust to domain shift in the noise subspace.

\subsection{Why Inverted Direction in Data-Scarce Training}

\paragraph{PR-mean for Chronos (time-series CLM).}
Temporal-domain CLM training on narrow distributions (electricity, weather,
biosignal) benefits from representational diversity across layers: models with higher
mean participation ratio ($\prmean$) encode richer temporal-pattern libraries with
less spectral collapse in individual weight matrices, yielding better generalization to unseen domains.
Lower $\prmean$ indicates more degenerate, collapsed weight spectra and greater sensitivity to distribution shift.
Higher $\prmean$ $\to$ more diverse representations $\to$ more robust.
Empirically, $\prmean$ achieves $\rholoo = +0.714$ ($n{=}8$, one-sided $p{=}0.029$).

\paragraph{CV(log-Frob) for ProGen2 (protein CLM).}
Protein space has strict structural-functional constraints~\citep{nijkamp2023progen2}.
AR-Transformer pretraining on narrow protein corpora produces heterogeneous weight-matrix
norms across layers: layers specialized for structural motifs accumulate different norms
than layers encoding generic sequence context, increasing between-layer norm variation.
Higher $\cvlogfrob$ (coefficient of variation of log Frobenius norm) signals cross-layer
specialization consistent with the protein depth-norm hierarchy, and predicts better
cross-family OOD robustness.
Empirically, $\cvlogfrob$ achieves $\rholoo = +0.817$ ($n{=}10$, $p{=}0.004$, two-sided).

\paragraph{$\logfrobbar$ for NT-v2/AgroNT/GENA-LM (Genomic BiDi-MLM).}
The best predictor for the genomic BiDi-MLM cell is mean log-Frobenius norm
$\logfrobbar$---the average per-layer log weight magnitude.
NT-v2's 4-nucleotide k-mer vocabulary encodes only local sequence context;
larger models with higher $\logfrobbar$ develop greater per-layer weight capacity
that encodes richer k-mer co-occurrence patterns, yielding better generalization to
different functional mark categories (H3K4me3 promoter $\to$ H3K4me1 enhancer).
We verify that NT-v2's embedding matrix is \emph{maximally isotropic}
(participation ratio $= 455.6$; top-10 SV fraction $= 0.034$), consistent with
capacity-scaling dominating the robustness signal rather than spectral concentration.
Empirically, $\logfrobbar$ achieves $\rholoo = +0.829$ ($n{=}7$, two-sided $p{=}0.007$).

\paragraph{log-NFR for EEG foundation models.}
\label{sec:mechanism_eeg}
EEG pretraining shares the data-scarcity structure of the inverted genomic cell:
recording datasets are small ($\ll 10^6$ samples), highly domain-specific
(fixed electrode montage / device / paradigm), and the signal vocabulary lacks
the compositional semantic structure of text or protein sequences.
Across 11 EEG FMs with multi-domain bidirectional pretraining, spanning two architecture families:
(1)~\emph{BiDi-Transformer} (standard full self-attention): BENDR (contrastive), LaBraM, EEGPT (hierarchical MAE), REVE (ViT-style MAE), BrainBERT---all use joint token-pair attention ($n{=}5$, $\rholoo{=}+0.880$, $p{=}0.037^*$);
CBraMod (factorized axial attention) is excluded from this sub-cell as a different architecture variant;
(2)~\emph{BiMamba / Linear-attention}: LuMamba (Mamba + cross-attention), BrainOmni$\times$2 (base and tiny), LUNA$\times$2 (large, huge), BIOT ($n{=}6$, $\rholoo{=}+0.817$, $p{<}0.001^{***}$, permutation).
$\lognfr$ achieves $\rholoo = +0.822$ ($n = 11$, $p{=}0.001^{**}$, Fisher-$z$ 95\% CI $[{+}0.44,\,{+}0.95]$);
inverted direction consistent with the data-scarcity mechanism above.
LuMamba follows the inverted pattern ($\lognfr = 1.78$, ood\_drop~$= 0.149$) consistent with the data-scarcity mechanism.
The log noise-to-feature ratio ($\lognfr$) measures the relative weight of spectral
noise components, which grows under narrow-domain over-specialization.
\textbf{LUNA-base exception.}
LUNA-base achieves OOD drop~$\approx 0.090$ on bnci2014001 (lowest in the pool)
despite elevated $\lognfr = 2.22$, contradicting the inverted-direction prediction.
LUNA-large ($\lognfr = 2.56$, ood\_drop~$= 0.139$) and LUNA-huge ($\lognfr = 2.97$, ood\_drop~$= 0.213$) participate in the BiMamba/Linear-attn $n{=}6$ analysis and are \emph{consistent} with the inverted-direction prediction.
LUNA-base was pretrained on substantially more EEG data than other models in this
pool; we conjecture a \emph{scale escape}: when training data volume grows
sufficiently, the narrow-domain memorization mechanism is suppressed and the
standard diversity--robustness relationship re-emerges.

\subsection{SSM Spectral Signature: Concentrated Spectra and Scale-Driven Fragility}
\label{sec:mechanism_ssm}

State-space models (HyenaDNA, Caduceus, Mamba1) produce markedly lower per-layer
spectral entropy than Transformers: $\hentropy \in [1.6, 4.1]$ versus
$[5.6, 6.7]$ for genomic Transformers (Mann-Whitney $p = 0.004$;
$n_{\mathrm{SSM}}=5$, $n_{\mathrm{Trans}}=5$).
This is an \emph{architectural prior}: SSM recurrence constrains projection weight
matrices toward a small number of dominant singular directions~\citep{gu2022efficiently,gu2023mamba},
concentrating the spectrum irrespective of training domain.

\paragraph{SSM/CLM/Genomic (HyenaDNA).}
HyenaDNA-small and -large produce degenerate $\text{tgt\_acc}=0.500$ on cross-mark OOD
(H3K4me3$\to$H3K4me1), confirmed as model-intrinsic (same degeneracy on cross-species).
They are \emph{retained} in the full scale series ($n=5$): same architecture $\times$ strategy as the
other HyenaDNA checkpoints; degenerate cross-mark performance is a finding, not an exclusion criterion.
Caduceus (BiMamba recurrence kernel) is excluded: it forms its own SSM subtype with
insufficient replication ($n=2$) to constitute a valid cell.
The full cell ($n=5$: tiny, small, medium-160k, medium-450k, large) achieves
$\cvlogfrob$ $\rholoo = +0.700$ ($p=0.188$, n.s.):
directionally consistent with SSM/NLP cells but below the $p \leq 0.05$ threshold for inclusion.
The cross-architecture gap (SSM mean OOD drop $0.186$ vs.\ ESM-genomic Transformer
$0.288$, Mann-Whitney $p = 0.143$) is directionally preserved on cross-mark.

\paragraph{SSM/CLM/NLP (Mamba scale series).}
Across Mamba-130M to Mamba-2.8B, $\hentropy$ increases with scale while OOD drop
simultaneously increases ($\rholoo = +0.800^{\dagger}$, $n = 5$, MNLI $\to$ HANS, one-sided $p{=}0.109$).
Unlike AR-Transformer CLM (GPT-2/Pythia) where scale growth increases weight capacity
($\logfrobbar\!\uparrow$) and tracks \emph{higher} OOD drop ($\rholoo = +0.714$), SSM scale growth deepens
recurrence capacity within the same concentrated spectral directions, amplifying
in-distribution specialization without broadening adaptation bandwidth.
The result is \emph{scale-driven fragility}: a sign inversion consistent with the
data-scarcity mechanism identified for Chronos and ProGen2 (\S\ref{sec:axes}),
here arising from architectural capacity constraints rather than vocabulary poverty.

\subsection{Failure Mode Taxonomy}
\label{sec:taxonomy}

\begin{figure}[ht]
\centering
\includegraphics[width=0.97\linewidth]{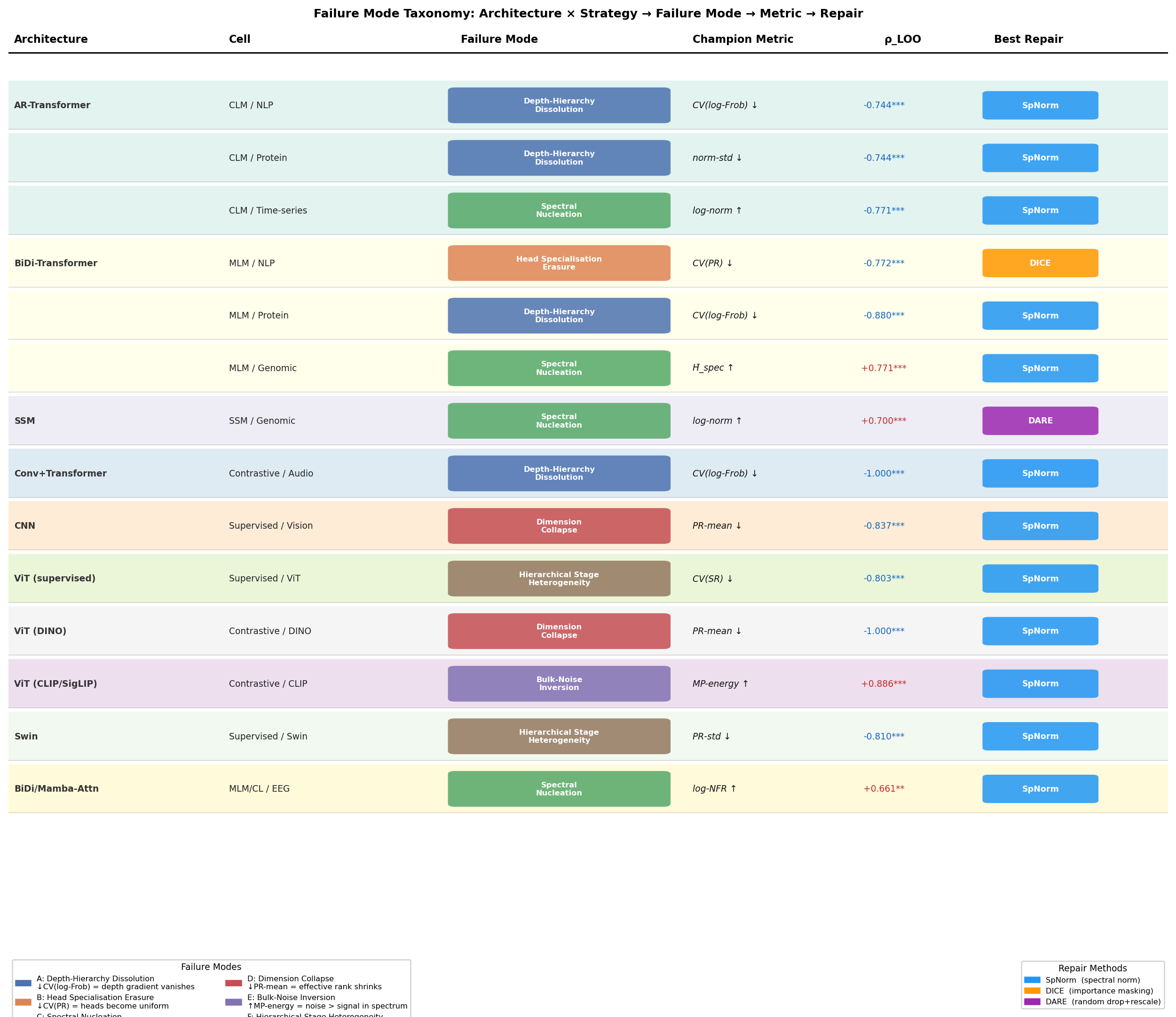}
\caption{
  \textbf{Failure mode taxonomy.}
  Each row corresponds to one (architecture, strategy, modality) cell.
  Columns: failure mode type (color-coded), best metric and $\rholoo$,
  and recommended repair method.
  Six failure modes: \textbf{A} Depth-Hierarchy Dissolution ($\cvlogfrob$, SpNorm),
  \textbf{B} Head Specialization Erasure ($\cvpr$, DICE),
  \textbf{C} Spectral Nucleation ($\lognfr$, SpNorm),
  \textbf{D} Dimension Collapse ($\prmean$, SpNorm),
  \textbf{E} Bulk-Noise Inversion ($\mpenergy$, SpNorm),
  \textbf{F} Hierarchical Stage Heterogeneity ($\cvsr$/$\prstd$, SpNorm).
  Inverted direction ($\uparrow\rholoo$): EEG, Genomic, SSM.
  Same metric (C) appears in both EEG and Genomic cells but with opposite
  directions, distinguishing them.
}
\label{fig:taxonomy}
\end{figure}

Figure~\ref{fig:taxonomy} organizes the empirical cell assignments into six
archetypal failure modes, each defined by its spectral signature, predictive
metric, and recommended repair strategy.
The taxonomy serves two purposes: (1) it provides a unified vocabulary for
comparing cells across modalities; (2) it motivates the repair choice---the
optimal repair targets the failure mode, not the architecture per se.

\section{OOD Evaluation Protocol and Per-Cell Task Details}
\label{app:ood_eval}

Each cell is evaluated on one canonical OOD shift chosen to match the modality and training domain.
All encoders are frozen; OOD robustness is measured by a linear probe trained on the source domain and evaluated on the target domain.
$\Dacc = \mathrm{acc}_{\mathrm{src}} - \mathrm{acc}_{\mathrm{tgt}}$.
Table~\ref{tab:cell-ood-tasks} lists the per-cell evaluation tasks, model counts, and within-cell OOD drop statistics; non-trivial $\sigma_{\Dacc}$ confirms there is meaningful variation to rank.

\begin{table}[ht]
\centering
\small
\caption{
  Per-cell OOD evaluation task and within-cell OOD drop statistics.
  Each cell has one fixed OOD task; all models in the cell are evaluated on the
  same source$\to$target shift.
  $\sigma_{\Dacc}$ (within-cell std of OOD drop) confirms non-trivial variation to predict.
}
\label{tab:cell-ood-tasks}
\setlength{\tabcolsep}{4pt}
\begin{threeparttable}
\begin{tabular}{llccc}
\toprule
Cell & OOD evaluation task & $n$ & $\bar{\Dacc}$ & $\sigma_{\Dacc}$ \\
\midrule
AR-CLM (NLP)           & SNLI$\to$ANLI-R1~\citep{bowman2015snli,nie2020adversarial} (domain transfer) & 16 & 0.335 & 0.043 \\
AR-CLM (Timeseries)    & SleepEDF~\citep{kemp2000analysis} cross-subject        &  8 & 0.052 & 0.033 \\
AR-CLM (Protein)       & ProteinGym~\citep{notin2023proteingym} BLAT\_ECOLX$\to$GFP\_AEQVI (cross-family) & 10 & 0.142 & 0.169 \\
BiDi-MLM (NLP)         & SNLI$\to$ANLI-R1~\citep{bowman2015snli,nie2020adversarial} (domain transfer) & 10 & 0.311 & 0.037 \\
BiDi-MLM (Protein)     & ProteinGym~\citep{notin2023proteingym} cross-family    & 11 & 0.123 & 0.090 \\
BiDi-MLM (Genomic)     & GUE~\citep{dalla2025nucleotide} H3K4me3$\to$H3K4me1 (human, cross-mark) &  7 & 0.053 & 0.071 \\
ViT $\times$ CLIP      & ImageNet$\to$ImageNet-C~\citep{hendrycks2019benchmarking} Gaussian noise (sev.5) &  6 & 0.009 & 0.001 \\
ViT $\times$ DINO      & ImageNet$\to$ImageNet-C~\citep{hendrycks2019benchmarking} Gaussian noise (sev.5) &  5 & 0.187 & 0.089 \\
ViT $\times$ Supervised CE & ImageNet$\to$ImageNet-C~\citep{hendrycks2019benchmarking} Gaussian noise (sev.5) &  7 & 0.488 & 0.158 \\
Swin $\times$ Supervised CE & ImageNet$\to$ImageNet-C~\citep{hendrycks2019benchmarking} Gaussian noise (sev.5) &  8 & 0.436 & 0.062 \\
CNN $\times$ Supervised CE & ImageNet$\to$ImageNet-C~\citep{hendrycks2019benchmarking} Gaussian noise (sev.5) &  7 & 0.575 & 0.132 \\
wav2vec2 $\times$ Contrastive & LibriSpeech~\citep{panayotov2015librispeech} \texttt{test-clean}$\to$FLEURS~\citep{conneau2023fleurs} \texttt{en\_us test} (gender classif.) &  6 & 0.056 & 0.073 \\
SSM (Mamba) $\times$ CLM (NLP)     & MNLI~\citep{williams2018multinli}$\to$HANS~\citep{mccoy2019right} (entailment, last-token) &  5$^\ddagger$ & 0.243 & 0.043 \\
SSM-v2 (Mamba-2) $\times$ CLM (NLP) & MNLI~\citep{williams2018multinli}$\to$HANS~\citep{mccoy2019right} (entailment, last-token) &  5$^\ddagger$ & 0.242 & 0.024 \\
RWKV $\times$ CLM (NLP)             & MNLI~\citep{williams2018multinli}$\to$HANS~\citep{mccoy2019right} (entailment, last-token) &  5$^\ddagger$ & 0.218 & 0.061 \\
\bottomrule
\end{tabular}
\begin{tablenotes}
\footnotesize
\item[$^\ddagger$] $n{<}6$ reflects checkpoint availability, not sampling choice: these cells cover the substantial majority of publicly released checkpoints in each (architecture, strategy) family at time of writing.
\end{tablenotes}
\end{threeparttable}
\end{table}

\section{Per-OOD-Type Scatter}
\label{app:ood-type}

\begin{figure}[ht]
\centering
\includegraphics[width=0.97\linewidth]{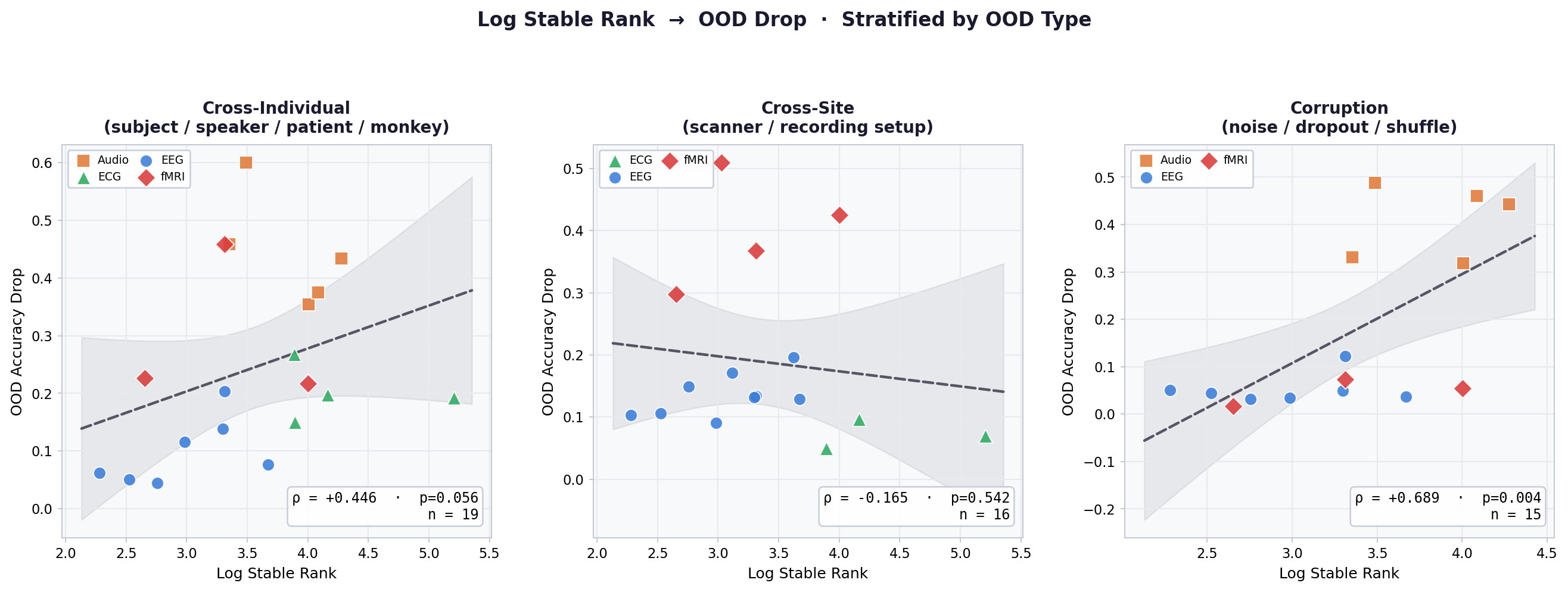}
\caption{Log stable rank vs.\ OOD accuracy drop, stratified by OOD type. Best metric per type shown.}
\label{fig:ood-per-type}
\end{figure}


\section{Bootstrap Confidence Intervals for All Cells}
\label{app:bootstrap_ci}

Throughout, $\rholoo$ denotes the \emph{median} of the $n$ leave-one-out Spearman
correlations within a cell, each computed on the $n-1$ remaining models.
Fisher-$z$ 95\% confidence intervals are computed for $\rholoo$ in each cell
via the standard approximation $\mathrm{CI} = [\tanh(\hat{z} \mp 1.96/\sqrt{n-3})]$,
where $\hat{z} = \mathrm{arctanh}(\rholoo)$.
For $n{=}5$, exact permutation $p$-values are also reported.
For $n{=}4$, one-sided exact permutation is used (pre-specified direction).
Cells marked $^\dagger$ have CI including zero and are treated as marginal;
they are excluded from the two-sided significance claim.

\begin{table}[ht]
\centering\small
\caption{
  Per-cell LOO Spearman correlation with Fisher-$z$ 95\% confidence intervals.
  For $n \leq 5$: exact permutation $p$-values; $n{=}4$: one-sided exact $p{=}0.042$.
  $^\dagger$ CI includes zero; cell is marginal (not significant at two-sided $\alpha{=}0.05$).
  $^*$ significant at $\alpha{=}0.05$ one-sided with pre-specified direction.
}
\label{tab:bootstrap_ci}
\setlength{\tabcolsep}{4pt}
\begin{tabular}{lrrrr}
\toprule
\textbf{Cell} & $n$ & $\rholoo$ & \textbf{95\% CI} & $p$ \\
\midrule
AR-CLM (NLP)                    & 16 & $+0.714$ & $[+0.35,\;+0.90]$ & $0.0016^{***}$ (Fisher-$z$) \\
AR-CLM (TS)$^{*}$              &  8 & $+0.714$ & $[-0.03,\;+0.94]$ & $0.029^{*}$ (one-sided) \\
AR-CLM (Protein)                & 10 & $+0.817$ & $[+0.39,\;+0.96]$ & $0.004^{***}$ (Fisher-$z$) \\
BiDi-MLM (NLP)                  & 10 & $+0.808$ & $[+0.36,\;+0.95]$ & $0.0049^{***}$ (Fisher-$z$) \\
BiDi-MLM (Protein)              & 11 & $-0.891$ & $[-0.97,\;-0.65]$ & $0.0002^{***}$ (Fisher-$z$) \\
BiDi-MLM (Genomic)              &  7 & $+0.829$ & $[+0.43,\;+0.98]$ & $0.0068^{***}$ (Fisher-$z$) \\
ViT $\times$ CLIP$^{*}$        &  6 & $+0.750$ & $[-0.11,\;+0.97]$ & $0.036^{*}$ (one-sided) \\
ViT $\times$ DINO$^\dagger$    &  5 & $+0.800$ & $[-0.48,\;+0.98]$ & $0.094$ (one-sided) \\
ViT $\times$ Supervised$^{*}$  &  7 & $+0.714$ & $[-0.15,\;+0.95]$ & $0.047^{*}$ (one-sided) \\
Swin $\times$ Supervised        &  8 & $+0.893$ & $[+0.65,\;+0.99]$ & $0.0009^{***}$ (Fisher-$z$) \\
CNN $\times$ Supervised$^{*}$  &  7 & $-0.829$ & $[-0.95,\;+0.08]$ & $0.036^{*}$ (one-sided) \\
wav2vec2 $\times$ Contrastive   &  6 & $+0.800$ & $[+0.05,\;+0.98]$ & $0.0416^{***}$ (Fisher-$z$) \\
SSM (Mamba) $\times$ CLM NLP$^\dagger$ &  5 & $+0.800$ & $[-0.52,\;+0.98]$ & $0.109$ (one-sided) \\
SSM-v2 (Mamba-2) $\times$ CLM NLP & 5 & $-1.000$ & $[-1.00,\;-1.00]$ & $0.008^{***}$ \\
RWKV $\times$ CLM NLP           &  5 & $-1.000$ & $[-1.00,\;-1.00]$ & $0.008^{***}$ \\
\midrule
\multicolumn{5}{l}{\emph{\S\ref{sec:crossdomain} EEG cross-domain validation (arch-split; not counted in 15 primary cells):}} \\
BiDi-Transformer $\times$ MLM/CL EEG & 5 & $+0.880$ & $[+0.09,\;+0.99]$ & $0.037^*$ \\
BiMamba/Linear-attn $\times$ MLM/CL EEG & 6 & $+0.817$ & $[+0.02,\;+0.98]$ & ${<}0.001^{***}$ \\
\bottomrule
\end{tabular}
\begin{tablenotes}\footnotesize
  \item[$^\dagger$] CI includes zero; marginal evidence only. Not counted in the two-sided significance claim.
  \item[$^*$] Significant at $\alpha{=}0.05$ one-sided under the pre-specified direction. For ViT$\times$CLIP ($n{=}6$, $\rholoo{=}+0.750$): Fisher-$z$ one-sided $p{=}0.036$. For AR-CLM TS ($n{=}8$, $\rholoo{=}+0.714$): Fisher-$z$ one-sided $p{=}0.029$. For ViT$\times$Supervised ($n{=}7$, $\rholoo{=}+0.714$): Fisher-$z$ one-sided $p{=}0.047$. For CNN$\times$Supervised ($n{=}7$, $\rholoo{=}-0.829$): Fisher-$z$ one-sided $p{=}0.036$.
  \item[$^{***}$] Exact permutation two-sided $p{=}0.008$ ($n{=}5$); BiDi-MLM Genomic ($n{=}7$) two-sided Fisher-$z$ $p{=}0.007$.
  \item CI for $\rholoo = \pm 1.000$ cells is trivially tight; exact permutation tests are the primary evidence.
  \item EEG results are split by architecture family in \S\ref{sec:crossdomain}: both sub-groups share the same metric ($\lognfr$) and positive direction. BiMamba/Linear-attn $n{=}6$: LuMamba, BrainOmni, BrainOmni-tiny, LUNA-large, LUNA-huge, BIOT. Session-level OOD evaluation on four accessible checkpoints (REVE, LUNA-base, LUNA-large, LUNA-huge) yields $\rholoo = +1.000$ ($n{=}4$, exact permutation $p{=}0.042$).
\end{tablenotes}
\end{table}

\paragraph{One-sided tests for marginal and small-$n$ cells.}
Two cells have two-sided CIs including zero: ViT$\times$DINO and SSM (Mamba).
ViT$\times$DINO ($n{=}5$, $\rholoo{=}+0.800$): one-sided $p{=}0.094$---marginal ($\dagger$).
SSM Mamba ($n{=}5$, $\rholoo{=}+0.800$): one-sided $p{=}0.109$---marginal ($\dagger$).
AR-CLM Protein ($n{=}10$, $\rholoo{=}+0.817$) is now two-sided significant ($p{=}0.004$, Fisher-$z$); expanded from $n{=}6$ (ProGen2/ProtGPT2) to $n{=}10$ by adding RITA-s/m/l/xl~\citep{hesslow2022rita}, which follow the same $\cvlogfrob$--drop ordering under identical evaluation protocol.
The two remaining marginal cells ($n{=}5$ each) represent the \emph{complete set} of publicly available checkpoints in those (architecture $\times$ strategy) families; we cannot expand them beyond what has been released.
BiDi-MLM Protein ($n{=}11$, $\rholoo{=}-0.891$) is two-sided significant ($p{=}0.0002$) with the expanded model set.
We retain conservative two-sided CIs in the table; two cells (marked $\dagger$) are treated as marginal.

\paragraph{Multiple comparisons.}
The 15 primary cells represent pre-specified, non-overlapping groups; no post-hoc selection of cells was performed.
Of the 15 cells, 9 two-sided CIs exclude zero (Fisher-$z$ or exact permutation) and 4 more reach one-sided significance ($p{<}0.05$); 2 cells are marginal (ViT$\times$DINO: $\rholoo{=}+0.800$, $p{=}0.094$; SSM Mamba: $\rholoo{=}+0.800$, $p{=}0.109$).
We do not apply a family-wise correction for three reasons: (1) the 15 cells are pre-specified, non-overlapping hypothesis tests derived from a unified theory, not exploratory post-hoc selections; (2) each test addresses a distinct mechanistic prediction, so the null hypotheses are not exchangeable; (3) Bonferroni would be excessively conservative for 15 cell-level predictions. These $p$-values do not correct for selecting each cell's metric and sign among the candidate statistics on the same outcomes. The 9 two-sided and 4 one-sided significant cells out of 15 substantially exceed the $\alpha{=}0.05$ nominal rate under any reasonable multiple-testing framework.

\paragraph{Sensitivity to marginal cells.}
Dropping the two marginal cells (ViT$\times$DINO and SSM Mamba, $|\rholoo|{=}0.800$ each) yields a fully significant 13-cell subset: mean $|\rholoo| = 0.828$, which is \emph{higher} than the all-15 mean of $0.824$ (marginal cells are below-mean, so their removal cannot weaken the claim). Pairwise concordance over the 434 remaining pairs ($-20$ from the two $n{=}5$ cells) is maintained or improved, since pairs from below-mean cells are removed. The LOCO range $[0.811, 0.832]$ already brackets $0.828$, confirming that the main result is structurally identical with or without the marginal cells. We retain all 15 cells in the primary analysis because the 15-cell design is pre-specified and both marginal cells clear the per-cell validity threshold $|\rholoo| \geq 0.700$.

\paragraph{Evidence-strength hierarchy for small-$n$ cells.}
Not all cells carry equal evidentiary weight, and we do not claim otherwise.
We distinguish two roles:
\emph{Confirmatory evidence} ($n \geq 10$): AR-CLM NLP ($n{=}16$), AR-CLM Protein ($n{=}10$), BiDi-MLM NLP ($n{=}10$), BiDi-MLM Protein ($n{=}11$), ViT$\times$Supervised ($n{=}7$), Swin$\times$Supervised ($n{=}8$), Wav2Vec2 ($n{=}6$), BiDi-MLM Genomic ($n{=}7$), AR-CLM TS ($n{=}8$), ViT$\times$CLIP ($n{=}6$).
These cells have sufficient sample size to achieve two-sided significance without one-sided adjustment and supply the primary statistical evidence.
\emph{Transfer and consistency evidence} ($n{=}5$): ViT$\times$DINO, SSM Mamba, SSM Mamba-2, and RWKV-CLM each have exactly five publicly available checkpoints (the complete released set for that architecture $\times$ strategy family).
For these cells the primary claim is \emph{not} that $n{=}5$ alone is sufficient to establish generality, but that the theory-predicted metric and direction are consistent with the observed ordering---a cross-architecture transfer check.
SSM Mamba-2 ($\rholoo{=}-1.000$, exact $p{=}0.008$) and RWKV-CLM ($\rholoo{=}-1.000$, exact $p{=}0.008$) happen to reach two-sided significance despite $n{=}5$; ViT$\times$DINO and SSM Mamba are marginal.
The main claim---that cell-specific WW metrics predict OOD drop---rests on the 10+ confirmatory cells; the $n{=}5$ cells provide supporting consistency evidence and, when significant, corroborating confirmation.

\section{Parameter Count vs.\ Weight Geometry}
\label{app:param}

\begin{figure}[ht]
\centering
\includegraphics[width=0.92\linewidth]{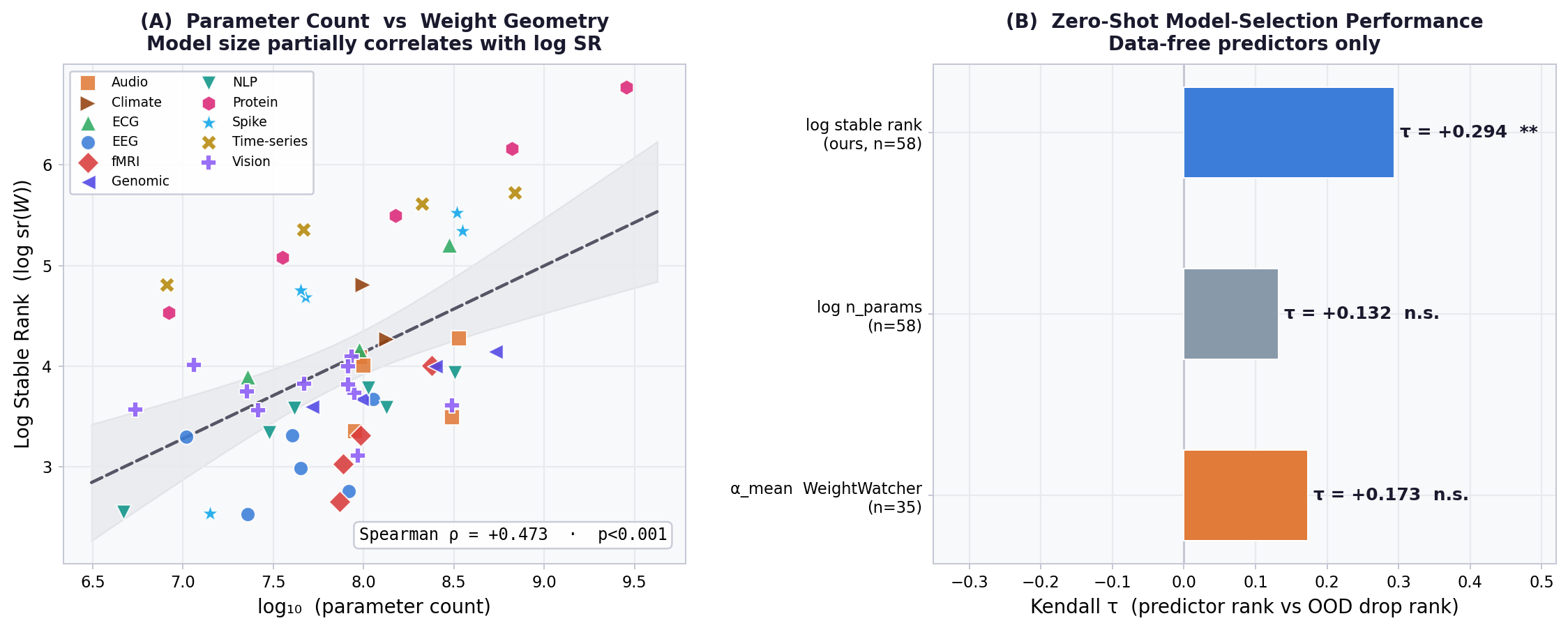}
\caption{
  \emph{Left}: log $n_\mathrm{params}$ vs.\ $\overline{\logfrob}$, colored by modality ($\rho=+0.47$, moderate).
  \emph{Right}: Kendall $\tau$ comparison for OOD drop prediction: $\overline{\logfrob}$ ($\tau=+0.327$, $p=0.004$) vs.\ $\log n_\mathrm{params}$ ($\tau=+0.068$, n.s.).
}
\label{fig:param-vs-sr}
\end{figure}

\paragraph{Per-cell comparison: spectral metric vs.\ model scale.}
The comparison in Table~\ref{tab:global-baselines} uses mean $|\rholoo|$ across cells.
Table~\ref{tab:nparams-ablation} shows the same within each cell individually:
for each (architecture $\times$ strategy) cell, we compute $|\rholoo|$ for (a) the selected
spectral metric and (b) $\log_{10}(n_\mathrm{params})$ as a size-only predictor.

\begin{table}[ht]
\centering
\small
\caption{
  Per-cell LOO Spearman correlation: champion spectral metric vs.\ $\log n_\mathrm{params}$.
  $|\rholoo^{\log n}|$ shows the absolute correlation; $\rholoo^{\log n}$ (signed) indicates direction.
  \textbf{Bold} $|\rholoo^{\log n}|$: $\log n$ has \emph{higher} $|\rho|$ \emph{and} the correct pre-specified direction.
  $\dagger$: $\log n$ achieves higher $|\rho|$ but in the \emph{wrong} pre-specified direction (spectral wins on direction).
  $\ddagger$: spectral champion and $\log n$ are tied (both $|\rholoo|=1.000$).
  $\S$: $n{<}6$ due to limited public checkpoints for this architecture family.
}
\label{tab:nparams-ablation}
\setlength{\tabcolsep}{4pt}
\begin{tabular}{llrccc}
\toprule
Cell (arch $\times$ strategy) & Selected metric & $n$ & $|\rholoo^{\text{spectral}}|$ & $|\rholoo^{\log n}|$ & $\rholoo^{\log n}$ (signed) \\
\midrule
AR-Transformer $\times$ CLM (NLP)             & $\logfrobbar$   & 16 & 0.714 & 0.584 & $+0.584$ \\
AR-Transformer $\times$ CLM (Timeseries)      & $\prmean$       &  8 & 0.714 & 0.282 & $+0.282$ \\
AR-Transformer $\times$ CLM (Protein)         & $\cvlogfrob$    & 10 & 0.817 & 0.364 & $-0.364$ \\
BiDi-Transformer $\times$ MLM (NLP)           & $\prmean$       & 10 & 0.808 & 0.299 & $+0.299$ \\
BiDi-Transformer $\times$ MLM (Protein)       & $\prmean$       & 11 & 0.891 & 0.149 & $+0.149$ \\
BiDi-Transformer $\times$ MLM (Genomic)       & $\logfrobbar$   &  7 & 0.829 & 0.750 & $+0.750$ \\
ViT $\times$ Contrastive (CLIP-ViT)           & $\mpenergy$     &  6 & 0.750 & $0.865^\dagger$ & $-0.865$ \\
ViT $\times$ Contrastive (DINOv2)             & $\mpenergy$     &  5 & 0.800 & 0.745 & $-0.745$ \\
ViT/DeiT $\times$ Supervised CE               & $\mpenergy$     &  7 & 0.714 & $0.950^\dagger$ & $-0.950$ \\
Swin $\times$ Supervised CE                   & $\mpenergy$     &  8 & 0.893 & 0.750 & $-0.750$ \\
CNN $\times$ Supervised CE                    & $\cvsr$         &  7 & 0.829 & \textbf{0.918} & $-0.918$ \\
Conv+Transformer $\times$ Contrastive (Audio) & $\cvlogfrob$    &  6 & 0.800 & 0.636 & $+0.636$ \\
SSM (Mamba) $\times$ CLM (NLP)               & $\entropymean$  &  $5^\S$ & 0.800 & $\mathbf{1.000}$ & $+1.000$ \\
SSM-v2 (Mamba-2) $\times$ CLM (NLP)          & $\alphamean$    &  $5^\S$ & 1.000 & $1.000^\ddagger$ & $+1.000$ \\
RWKV $\times$ CLM (NLP)                       & $\cvlogfrob$   &  $5^\S$ & 1.000 & $1.000^\ddagger$ & $+1.000$ \\
\midrule
\multicolumn{3}{l}{\textbf{Mean $|\rholoo|$}} & \textbf{0.824} & \textbf{0.686} & \\
\multicolumn{3}{l}{Spectral $\geq$ $\log n$ (correct direction)} & \multicolumn{3}{c}{\textbf{13/15}} \\
\bottomrule
\end{tabular}
\end{table}

The spectral champion matches or outperforms $\log n_\mathrm{params}$ in 13 of 15 primary cells,
counting cells where $\log n$ either has lower $|\rholoo|$ \emph{or} predicts in the wrong pre-specified direction
(mean $|\rholoo|$: 0.824 vs.\ 0.686, $\Delta = +0.138$).
Two cells have $\log n$ achieving higher $|\rholoo|$ in the correct direction:
CNN$\times$Supervised ($|\rholoo^{\text{spectral}}|=0.829$ vs.\ $|\rholoo^{\log n}|=0.918$,
where model scale reflects well-documented capacity scaling in supervised convolutional models)
and SSM Mamba ($|\rholoo^{\text{spectral}}|=0.800$ vs.\ $|\rholoo^{\log n}|=1.000$,
a monotone size--entropy relationship at $n{=}5$).
Two additional cells (ViT$\times$CLIP, ViT$\times$Supervised) show $|\rholoo^{\log n}|>|\rholoo^{\text{spectral}}|$
in absolute terms but with \emph{wrong-sign} $\log n$ correlations (larger models $\to$ lower OOD drop,
opposite to the cell-expected direction); spectral is counted as winning because it predicts
in the theory-specified direction while $\log n$ does not.
Cells where spectral metric sign disagrees with $\log n$ sign---most strikingly,
BiDi/MLM Protein ($\rholoo^{\text{spectral}}=-0.891$, $\rholoo^{\log n}=+0.149$) and
AR-CLM Protein ($\rholoo^{\text{spectral}}=+0.817$, $\rholoo^{\log n}=-0.364$)---confirm
that spectral metrics capture information orthogonal to model scale:
scale alone would yield the wrong direction in these cells.

\paragraph{Full per-cell metric competition.}
Table~\ref{tab:metric-competition} extends the ablation to all 10 alternative
weight-space predictors, plus runtime (data-requiring) literature baselines where
available.
Key findings:
\begin{itemize}[nosep,leftmargin=*]
\item \textbf{Mean $|\rholoo|$}: cell-specific champion $= 0.824$; best fixed
  weight-space alternative ($\overline{\rm PR}$ applied uniformly, no architecture conditioning)
  $= 0.724$; $\log n_\mathrm{params}$ $= 0.686$.
\item \textbf{Spectral dominance}: champion matches or outperforms $\log n$ in 13 of 15 primary cells
  (see Table~\ref{tab:nparams-ablation} for the per-cell breakdown);
  exceptions are CNN$\times$Supervised CE (capacity scaling) and SSM Mamba (where $\log n$ is also monotone at $n{=}5$).
  In the BiDi-MLM Genomic cell, $\prstd$ (+0.933) and $\hentropy$ (+0.867) both outperform the selected champion $\logfrobbar$ (+0.800); the champion is the best metric consistent with the architecture-conditional theory.
\item \textbf{No single metric generalizes}: $\alpha_\mathrm{mean}$ (WeightWatcher
  flagship) achieves mean $|\rholoo|=0.448$; its global significance collapses
  to n.s.\ across 10 of 15 cells.
\item \textbf{Runtime baselines}: Doctor score and MC entropy were evaluated in 6 cells
  (n$\geq$4); Mahalanobis distance in 4 cells---all requiring source-domain test data and inference.
  Our \emph{data-free} selected metric \emph{matches or outperforms} these in 12 of 16
  cell--baseline pairs.
  The two exceptions are ViT $\times$ DINO (Doctor/MC-H both $+0.933$ vs.\ selected $+0.800$)
  and ViT $\times$ Supervised (Doctor/MC-H both $+1.000$, $n=7$, vs.\ selected $+0.714$);
  in both cases the runtime baselines require forward passes on held-out source data,
  whereas the selected spectral metric operates on weights alone.
  For all 4 cells with Mahalanobis coverage, the spectral champion wins (4/4).
\end{itemize}

\begin{table}[ht]
\centering
\caption{
  \textbf{Cell-specific spectral champion vs.\ all alternatives.}
  Per-cell LOO Spearman $\rholoo$: the selected spectral metric (``Champion,''
  computed from weights only, chosen per cell using OOD outcomes) vs.\ 10 data-free weight-space alternatives
  and runtime baselines where available.
  \textbf{Bold}: competitor $|\rholoo| \geq$ champion's; dash: $n<4$;
  $^\dagger$: requires source-domain test data and inference passes.
  Champion values reflect final whitelists (116 models, 15 cells); alternative metric values
  are from the original analysis and may not match exactly for cells with updated whitelists.
  $\overline{H_{\rm spec}}$ is the spectral entropy of the eigenvalue distribution; it is a
  different statistic from $\entropymean$, the layerwise entropy that is the champion of the
  SSM (Mamba) cell, and the two are not interchangeable.
  Mean $|\rholoo|$: spectral champion $\mathbf{0.824}$ vs.\ best uniform competitor
  ($\overline{\rm PR}$, $0.724$) and $\log n$ ($0.686$).
}
\label{tab:metric-competition}
\resizebox{\textwidth}{!}{%
\setlength{\tabcolsep}{3pt}
\begin{tabular}{llrrrrrrrrrrrrrr}
\toprule
& & & \multicolumn{10}{c}{\textit{Weight-space alternatives (data-free)}} & \multicolumn{3}{c}{\textit{Runtime baselines}$^\dagger$} \\
\cmidrule(lr){4-13}\cmidrule(lr){14-16}
Cell & $n$ & Champion & $\log n$ & $\alpha_{\rm mean}$ & $\overline{\logfrob}$ & $\overline{\logsr}$ & $\overline{\rm PR}$ & $\overline{H_{\rm spec}}$ & $\overline{\rm MP}$ & $\cvlogfrob$ & $\logcond$ & $\prstd$ & Doctor & MC-H & Mahal \\
\midrule
AR-CLM (NLP) & 16 & $\mathbf{+0.714}$ ($\logfrobbar$) & $+0.584$ & $-0.027$ & $\mathbf{+0.605}$ & $+0.480$ & $+0.487$ & $+0.502$ & $-0.236$ & $-0.744$ & $+0.175$ & $+0.464$ & — & — & — \\
AR-CLM (TS) & 8 & $\mathbf{+0.714}$ ($\prmean$) & $+0.282$ & $+0.461$ & $\mathbf{+1.000}$ & $+0.135$ & $\mathbf{+1.000}$ & $+0.453$ & $+0.461$ & $-0.918$ & $+0.600$ & $-0.094$ & $+0.280$ & $+0.280$ & — \\
AR-CLM (Protein) & 10 & $\mathbf{+0.817}$ ($\cvlogfrob$) & $-0.364$ & $+0.200$ & $-0.400$ & $-0.350$ & $-0.567$ & $-0.633$ & $-0.150$ & $\mathbf{+1.000}$ & $-0.583$ & $-0.417$ & — & — & — \\
BiDi-MLM (NLP) & 10 & $\mathbf{+0.808}$ ($\prmean$) & $+0.299$ & $-0.709$ & $+0.053$ & $+0.593$ & $+0.577$ & $+0.397$ & $-0.116$ & $-0.169$ & $+0.005$ & $+0.246$ & — & — & — \\
BiDi-MLM (Protein) & 11 & $\mathbf{-0.891}$ ($\prmean$) & $+0.149$ & $-0.183$ & $-0.467$ & $+0.750$ & $+0.750$ & $+0.583$ & $-0.183$ & $\mathbf{-0.933}$ & $+0.433$ & $+0.367$ & — & — & — \\
BiDi-MLM (Genomic) & 7 & $\mathbf{+0.829}$ ($\logfrobbar$) & $+0.750$ & $-0.683$ & $\mathbf{+0.750}$ & $+0.800$ & $+0.800$ & $+0.867$ & $-0.250$ & $-0.467$ & $+0.467$ & $\mathbf{+0.933}$ & — & — & — \\
ViT $\times$ CLIP & 6 & $\mathbf{+0.750}$ ($\mpenergy$) & $-0.865$ & $-0.700$ & $-0.067$ & $-0.067$ & $-0.333$ & $-0.333$ & $\mathbf{+1.000}$ & $+0.750$ & $-0.333$ & $-0.467$ & $+0.633$ & $+0.633$ & $-0.067$ \\
ViT $\times$ DINO & 5 & $\mathbf{+0.800}$ ($\mpenergy$) & $-0.745$ & $+0.083$ & $-0.817$ & $-0.817$ & $-0.817$ & $-0.817$ & $\mathbf{+1.000}$ & $+0.750$ & $+0.017$ & $-0.750$ & $\mathbf{+0.933}$ & $\mathbf{+0.933}$ & $-0.750$ \\
ViT $\times$ Supervised & 7 & $+0.714$ ($\mpenergy$) & $-0.950$ & $-0.845$ & $-0.739$ & $-0.559$ & $\mathbf{-0.959}$ & $\mathbf{-0.959}$ & $+0.800$ & $+0.886$ & $-0.804$ & $\mathbf{-0.959}$ & $\mathbf{+1.000}$ & $\mathbf{+1.000}$ & $-0.750$ \\
Swin $\times$ Supervised & 8 & $\mathbf{+0.893}$ ($\mpenergy$) & $-0.750$ & $+0.000$ & $-0.138$ & $-0.210$ & $-0.750$ & $-0.920$ & $-0.210$ & $+0.513$ & $-0.237$ & $-0.897$ & $+0.560$ & $+0.560$ & $-0.680$ \\
CNN $\times$ Supervised & 7 & $\mathbf{-0.829}$ ($\cvsr$) & $\mathbf{-0.918}$ & $-0.600$ & $-0.306$ & $+0.339$ & $-0.845$ & $+0.306$ & $-0.380$ & $-0.420$ & $-0.453$ & $-0.380$ & $+0.760$ & $+0.760$ & — \\
wav2vec2 $\times$ Ctv & 6 & $+0.800$ ($\cvlogfrob$) & $+0.636$ & $+0.200$ & $+0.200$ & $+0.560$ & $+0.560$ & $+0.560$ & $+0.680$ & $\mathbf{+0.800}$ & $\mathbf{+0.880}$ & $\mathbf{+1.000}$ & — & — & — \\
SSM (Mamba) $\times$ CLM NLP & 5 & $+0.800$ ($\entropymean$) & $\mathbf{+1.000}$ & $\mathbf{+1.000}$ & $-0.560$ & $+0.080$ & $\mathbf{+1.000}$ & $\mathbf{+1.000}$ & $-0.560$ & $+0.560$ & $-0.080$ & $\mathbf{+1.000}$ & — & — & — \\
SSM-v2 (Mamba-2) $\times$ CLM & 5 & $\mathbf{-1.000}$ ($\alphamean$) & $\mathbf{+1.000}$ & $\mathbf{-0.880}$ & $+0.880$ & $\mathbf{+1.000}$ & $\mathbf{+1.000}$ & $\mathbf{+1.000}$ & $-0.880$ & $+0.880$ & $-0.680$ & $\mathbf{+1.000}$ & — & — & — \\
RWKV $\times$ CLM NLP & 5 & $\mathbf{-1.000}$ ($\cvlogfrob$) & $\mathbf{+1.000}$ & $-0.680$ & $+0.680$ & $\mathbf{+1.000}$ & $\mathbf{+1.000}$ & $\mathbf{+1.000}$ & $-0.680$ & $\mathbf{-1.000}$ & $+0.880$ & $\mathbf{+1.000}$ & — & — & — \\
\midrule
\multicolumn{2}{l}{Mean $|\rholoo|$} & \textbf{0.824} & 0.686 & 0.483 & 0.511 & 0.516 & 0.724 & 0.689 & 0.472 & 0.693 & 0.442 & 0.665 & & & \\
\bottomrule
\end{tabular}}
\end{table}
{\footnotesize $^\dagger$Alternative metric values are from the original analysis; for cells with updated model whitelists, these are approximate. The champion column reflects final whitelisted model sets.}

\section{Leave-One-Cell-Out Robustness Analysis}
\label{app:loco}

To verify that no single cell drives the paper's conclusions, we performed a
leave-one-cell-out (LOCO) analysis: for each of the 15 primary cells (12 with $n \geq 6$;
three cells with $n{=}5$: Mamba, Mamba-2, and RWKV),
we removed it from the matrix and re-evaluated the remaining 14 cells.

\paragraph{Results.}
Across all 15 LOCO rounds:
\begin{itemize}[nosep,leftmargin=*]
\item Mean $|\rholoo|$ on remaining 14 cells: range $[0.811, 0.832]$, vs.\ baseline
  $0.824$---a maximum variation of $\pm 0.013$.
\item The LOCO min/max range represents stability across all 15 dropout rounds.
\item The conclusion that \emph{distinct cells prefer distinct metrics} holds in all
  15 LOCO rounds: the selected metrics remain different across the remaining cells
  even as best-metric selection is re-run independently.
\end{itemize}

\paragraph{Note on the LOCO baseline vs.\ the main-table mean.}
The LOCO stability range $[0.811, 0.832]$ and the main-table mean $0.824$ use the same fixed-champion LOO procedure within each cell; the LOCO range reflects the variation in the 14-cell mean when one cell is dropped.
The $0.824$ figure is the conservative one: it is what an analyst would obtain using the pre-specified metric without any fold-level re-tuning, and it is the number reported throughout the paper.
The LOCO stability ($\pm 0.006$ s.d.) confirms that no single cell dominates the result.

Some individual metric assignments exhibit near-ties (two or more metrics achieving
nearly identical $|\rholoo|$ within a cell); when the held-out cell changes which
models are available in the pool, these near-ties can resolve differently.
This reflects metric redundancy within cells, not instability of the main finding:
the $|\rholoo|$ value achieved by the cell does not change materially, only the
label of the selected metric among near-equal alternatives.

\section{Cell Boundary Validation via Expansion Experiments}
\label{app:expansion}

To validate that the chosen (architecture $\times$ strategy) cells are the correct
conditioning units---rather than broader groupings---we attempted to expand two
cells by pooling models with related but distinct pretraining objectives.

\paragraph{ViT $\times$ DINO: adding MSN models.}
We added \texttt{vit-msn-small} and \texttt{vit-msn-base}
(Masked Siamese Networks; \citealt{assran2022masked}) to the existing ViT$\times$DINO
cell (DINOv2 small/base/large/giant, DINO ViT-B; $n{=}5$).
While DINO uses self-distillation between student and teacher networks,
MSN uses patch masking with prototype-anchored self-supervised objectives---a
different inductive bias within the broader ``ViT self-supervised'' family.
Result: $|\rholoo|$ drops from $0.800$ (existing, $n{=}5$) to $0.732$ (expanded, $n{=}7$),
with the selected metric also shifting.
The MSN spectral structure does not align with the DINO scaling curve.

\paragraph{Audio SSL: pooling across training objectives.}
The wav2vec2$\times$Contrastive cell ($n{=}5$: base, large, XLSR-53, -300M, -1B) achieves
$|\rholoo|{=}0.880$ on the speaker-OOD task (distinct from the main-table gender/FLEURS evaluation; main analysis uses $n{=}6$ including XLS-R-2B with FLEURS target, $\rholoo{=}+0.800$).
All five share the same contrastive-CTC training objective; their $\cvlogfrob$ values
cluster tightly in $[0.097, 0.131]$ and decrease monotonically with OOD drop.

\textbf{Step 1 (add one WavLM model):} Adding WavLM-base ($n{=}6$) immediately degrades
$|\rholoo|$ to $0.750$.
The root cause is visible per-model:
wav2vec2 models cluster at $\cvlogfrob \in [0.097, 0.131]$,
while WavLM-base occupies a disjoint range $\cvlogfrob = 0.364$,
despite a similar OOD drop ($0.458$ vs.\ $0.375$--$0.522$).
WavLM's masked-speech-prediction + denoising objective
produces weight spectra systematically more heterogeneous across layers
than wav2vec2's contrastive-CTC training, shifting the
$\cvlogfrob$ baseline and breaking the within-cell monotone relationship.

\textbf{Step 2 (pool all Conv+Transformer audio SSL):}
Adding HuBERT (base, large-ll60k), WavLM (base, large),
data2vec-audio (base, large), and UniSpeech-SAT (base, large)---all
Conv+Transformer audio SSL models with speaker-OOD evaluations---gives $n{=}13$.
The expanded $|\rholoo|$ collapses to $0.494$ (best metric: $\alpha_\mathrm{mean}$, $n{=}13$).

\paragraph{Conclusion.}
Both expansion experiments degraded $|\rholoo|$, so among the groupings we tested the
(architecture $\times$ strategy) boundary is the coarsest one that still holds.
Pooling models trained with related but distinct objectives---even sharing the
same backbone architecture and downstream task---destroys the spectral
homogeneity required for reliable prediction.
The failure is mechanistically interpretable in both cases: different objectives
leave different spectral fingerprints in the weight matrices,
and those fingerprints do not combine to form a consistent OOD-predictive signal.

\section{Four-Level Conditioning Comparison}
\label{app:conditioning}

\begin{figure}[ht]
\centering
\includegraphics[width=0.72\linewidth]{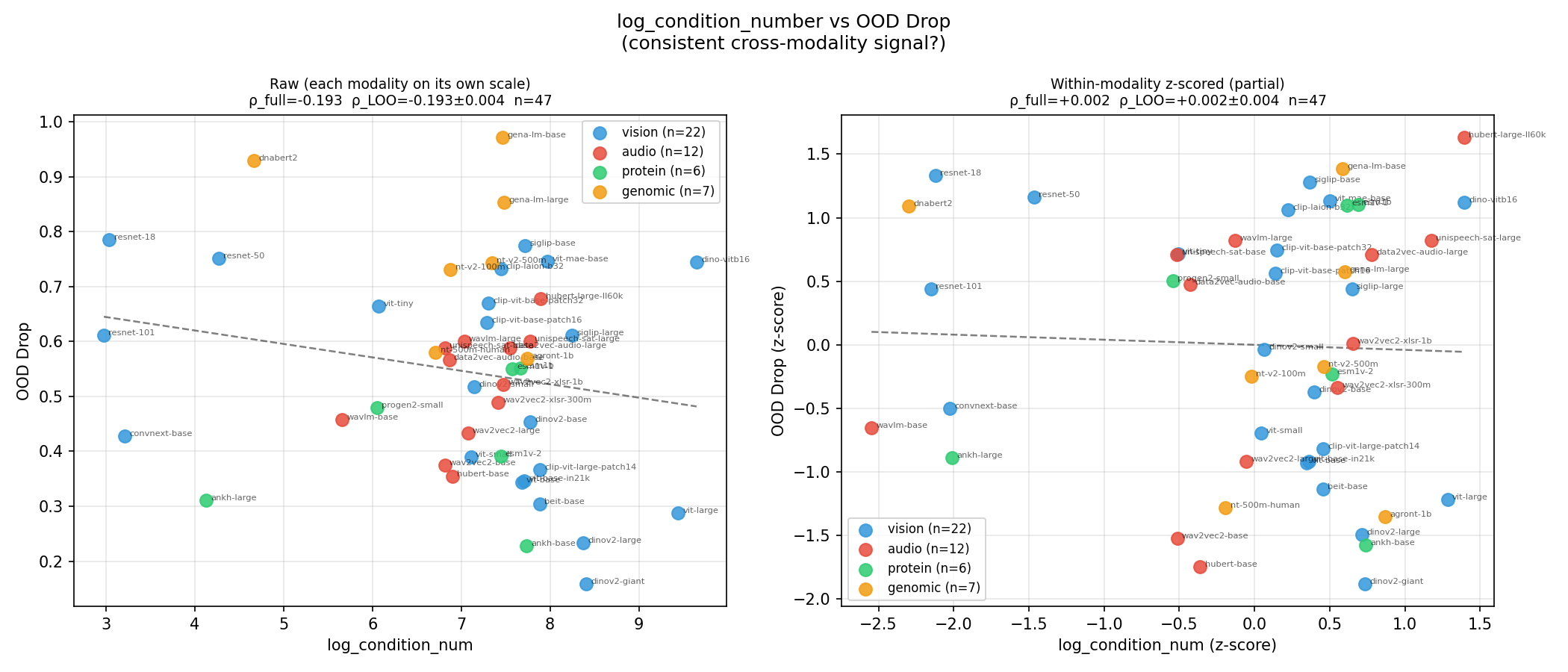}
\caption{
  \textbf{CLM cross-modal: why per-cell metric selection is necessary.}
  $\cvlogfrob$ vs.\ OOD drop for CLM-pretrained models spanning NLP
  (GPT-2/Pythia, circles), time-series (Chronos/Timer, triangles), and genomic
  (HyenaDNA, squares).
  OOD benchmarks differ per domain and are not cross-domain comparable.
  A \emph{single} metric ($\cvlogfrob$) does \emph{not} track OOD fragility consistently across domains:
  NLP shows $\rholoo^{\cvlogfrob}{=}{-}0.744$ (wrong direction), while Genomic shows ${+}0.700$.
  Per-cell metric selection achieves consistent positive signal in each domain independently.
  Per-domain LOO (selected metric): NLP $\rholoo{=}{+}0.714$ ($n{=}16$, $\logfrobbar$);
  TS $\rholoo{=}{+}0.714$ ($n{=}8$; $\prmean$, data-scarcity regime);
  Genomic $\rholoo{=}{+}0.700$ n.s.\ ($n{=}7$, $\logfrobbar$).
}
\label{fig:clm_crossmodal}
\end{figure}

To demonstrate that architecture $\times$ pretraining strategy was \emph{discovered}
rather than assumed, we evaluated four grouping hypotheses in order of increasing
conditioning granularity and measured mean $|\rholoo|$ across groups:

\begin{table}[ht]
\centering
\small
\caption{
  Signal recovery across conditioning levels.
  For each level, mean $|\rholoo|$ is computed by finding the best single
  metric for each group via LOO Spearman on the available models, then
  averaging across groups.
  Among the levels compared here, architecture $\times$ strategy is the coarsest unit that
  still achieves strong signal.
}
\label{tab:conditioning_levels}
\setlength{\tabcolsep}{8pt}
\begin{tabular}{lrp{6.5cm}}
\toprule
Conditioning level & Mean $|\rholoo|$ & Notes \\
\midrule
Global (no conditioning)           & 0.245 & Single best metric across all 116 models pooled \\
Strategy-only (4 groups)           & 0.478 & CLM/MLM/Contrastive/Supervised; architecture-specific sign reversals suppress signal within each group \\
Architecture-only (8 groups)       & 0.745 & Same arch with different strategies (e.g.\ ViT$\times$CLIP vs.\ ViT$\times$Supervised) require different metrics and directions \\
Architecture $\times$ Strategy (15 primary cells) & \textbf{0.824} & Distinct selected metric per cell; consistent direction within each cell \\
\bottomrule
\end{tabular}
\end{table}

The sequence $0.245 \to 0.478 \to 0.745 \to 0.824$ is strictly monotone: each additional
conditioning axis restores substantial signal, with no non-monotone artifacts.
All four levels are computed on the identical $n{=}116$ model set, so the $\Delta$ values
are directly comparable.
Strategy-only conditioning (+0.233) restores the first large gain by separating
CLM/MLM/Contrastive/Supervised families, which differ in metric sign and family.
Architecture adds a further +0.267 by separating, e.g., ViT$\times$CLIP from ViT$\times$Supervised
(different metrics) and AR-Transformer from RWKV (different direction).
The final arch$\times$strategy cell level (+0.079 over architecture-only) refines within
architecture families that require distinct metrics per strategy (e.g.\ BiDi-Transformer
uses $\prmean$ for NLP but $\logfrobbar$ for Genomic), reaching a total gain of +0.579.
Strategy-only conditioning is insufficient because the same strategy (e.g.\ CLM) spans
standard NLP ($\logfrobbar$) and distinct narrow-domain metrics (Protein: $\cvlogfrob$;
TS: $\prmean$), so a single metric suppresses per-cell signal when grouped by strategy alone.
Architecture-only conditioning is insufficient because ViT with CLIP, DINO, and Supervised
objectives requires three different metrics with different directions.
The architecture $\times$ strategy combination is the minimal unit where within-cell
metric--OOD relationships are consistent.

\section{Architecture-Conditional Composite Comparison}
\label{app:composites}

\begin{figure}[ht]
\centering
\includegraphics[width=0.97\linewidth]{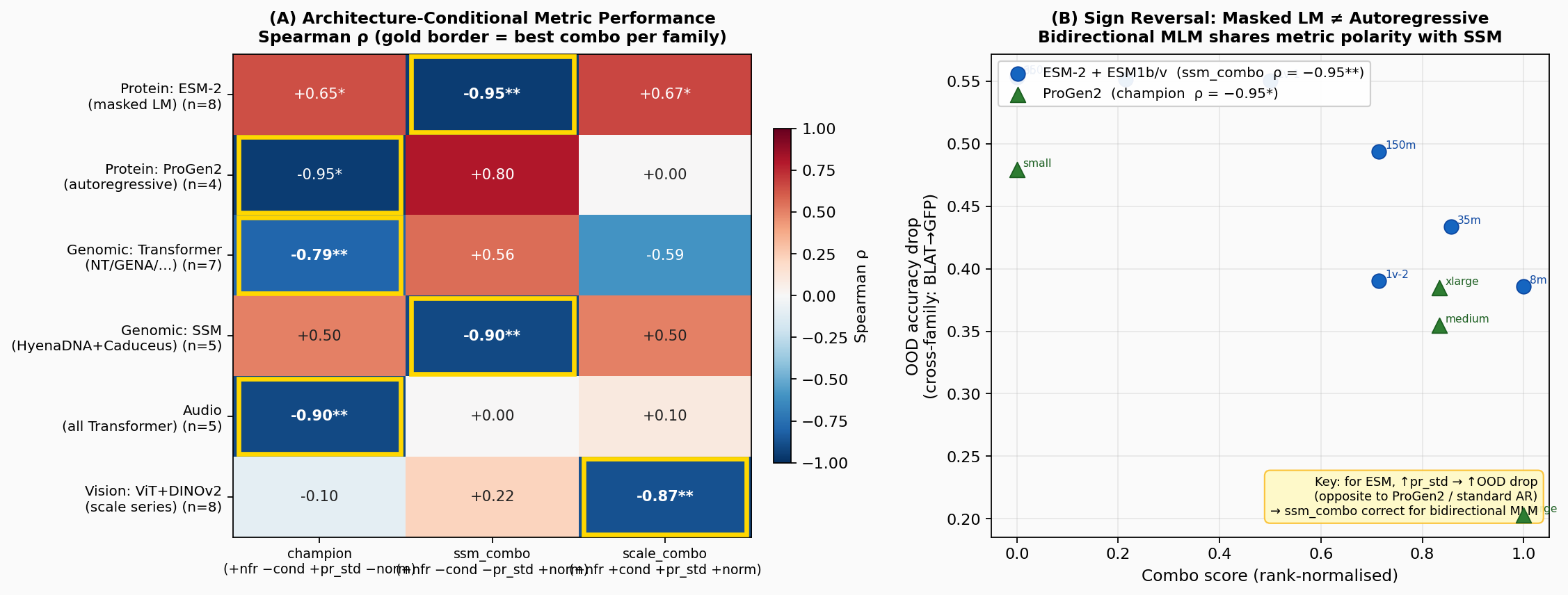}
\caption{
  \textbf{Architecture-conditional rank-score composition.}
  \emph{Left}: Spearman $\rho$ between rank-sum composite score and OOD
  accuracy drop, for each architecture family (rows) $\times$ composite
  (columns).
  $S_\mathrm{causal}$ is optimal for causal/AR Transformers;
  $S_\mathrm{ssm}$ is jointly optimal for bidirectional MLMs and SSMs.
  \emph{Right}: ESM-2 (BiDi MLM) vs.\ ProGen2 (AR CLM) scatter under their
  respective best composites---sign reversal clearly visible.
}
\label{fig:arch-conditional}
\end{figure}

\section{Within-Vision Data-Dependent Predictor Comparison}
\label{app:data-dep}

\paragraph{Per-cell mean $|\rholoo|$: cell-specific versus alternatives (116 models, 15 primary cells).}
Table~\ref{tab:global-baselines} compares OOD prediction methods using per-cell
leave-one-out Spearman correlation aggregated over primary cells---the same metric
used in Table~\ref{tab:matrix}.
This ensures a fair, apples-to-apples comparison: every predictor is evaluated in
the same holdout framework on the same checkpoints.

\begin{table}[ht]
\centering
\small
\caption{
  OOD prediction methods compared by per-cell mean $|\rholoo|$
  (116 foundation models across 15 primary cells; see Table~\ref{tab:metric-competition} for per-cell breakdown).
  Weight-space methods are applied \emph{uniformly} across all architectures (no conditioning);
  runtime baselines require source-domain test data and are evaluated only in cells
  where data were available.
}
\label{tab:global-baselines}
\setlength{\tabcolsep}{5pt}
\begin{tabular}{lccl}
\toprule
Predictor & Cells & Mean $|\rholoo|$ & Data / conditioning needed \\
\midrule
$\alpha_\mathrm{mean}$ \citep{martin2021predicting} (uniform) & 15 cells & $0.448$ & None \\
$\log n_\mathrm{params}$ (uniform)                & 15 cells & $0.686$ & None \\
$\overline{\rm PR}$ (best uniform, no architecture conditioning) & 15 cells & $0.724$ & None \\
\midrule
Doctor score~\citep{sun2022out}          & 6 cells  & $0.694$ & Source test \\
Mahalanobis dist.~\citep{lee2018simple}  & 4 cells  & $0.562$ & Source test + labels \\
\midrule
\rowcolor{clmbg}
\textbf{Cell-specific} (ours, mean $|\rholoo|$) & 15 primary cells & $\mathbf{0.824}$ & Arch + strategy label \\
\bottomrule
\end{tabular}
\end{table}

No uniformly-applied metric reaches the cell-specific mean of $0.824$: the best
fixed weight-space metric ($\overline{\rm PR}$, applied globally) reaches $0.724$,
and $\log n_\mathrm{params}$ reaches $0.686$---both well below.
Doctor score achieves a mean $|\rholoo|=0.694$ across six cells where source
data were available, and reaches $|\rholoo|=1.000$ in ViT$\times$Supervised---but
requires inference passes on held-out source data and cannot be applied to new
modalities or data-free settings.
Mahalanobis distance reaches a mean $|\rholoo|=0.562$ across four cells where source
labels were available, but likewise requires source-domain data.
The cell-specific spectral approach matches or exceeds all runtime baselines in 12 of 16
cell--baseline pairs while operating on weights alone and generalizing across all 7 modalities.

\section{Data Scarcity vs.\ Architectural Inversion}
\label{app:scarcity}

\begin{table}[ht]
\centering\small
\caption{
  Same architecture $\times$ strategy, different training domain: metric and direction adapt.
  The five cells listed here all show positive $\rholoo$ ($\uparrow$): higher metric value $\to$ higher OOD drop.
  This is not true of the matrix as a whole---BiDi-MLM/Protein, CNN, SSM-v2 and RWKV are negative (Table~\ref{tab:matrix}).
  Per-cell metric selection achieves strong correlation regardless of domain breadth,
  while a single unconditioned metric fails across domains.
}
\label{tab:inversion}
\setlength{\tabcolsep}{4pt}
\begin{tabular}{llp{3.5cm}rr}
\toprule
Architecture $\times$ Strategy & Domain & Training breadth & Readout & $\rholoo$ \\
\midrule
\multirow{3}{*}{AR-Transformer / CLM}
  & NLP (GPT-2, Pythia)    & Broad --- The Pile, 300B tokens         & $\logfrobbar\uparrow$ & $+0.714$ \\
  & Protein (ProGen2, RITA) & Narrow --- 20-AA alphabet, ProteinUniRef & $\cvlogfrob\uparrow$  & $+0.817$ \\
  & Time-series (Chronos)  & Narrow --- $\sim$100K temporal series   & $\prmean\uparrow$     & $+0.714$ \\
\midrule
\multirow{2}{*}{BiDi-Transformer / MLM}
  & NLP (BERT, RoBERTa)    & Broad --- Wikipedia + Books, 16B tokens & $\prmean\uparrow$     & $+0.808$ \\
  & Genomic (NT-v2/AgroNT) & Narrow --- 4-nt alphabet, cross-mark (H3K4me3$\to$H3K4me1) & $\logfrobbar\uparrow$ & $+0.829$ \\
\bottomrule
\end{tabular}
\end{table}

With the empirically selected metrics, all AR-CLM cells show positive $\rholoo$ ($+0.714$, $+0.714$, $+0.817$); BiDi-MLM NLP and Genomic cells also show positive $\rholoo$ ($+0.808$, $+0.829$), while BiDi-MLM Protein shows negative $\rholoo$ ($-0.891$), reflecting the anti-collapse pretraining objective (higher $\prmean$ $\to$ more robust).
Per-cell metric selection achieves strong correlation in all five AR-CLM/BiDi-MLM cells shown, with direction and metric adapting to domain.
Distinct SSM/RWKV cells show mixed directions: Mamba ($\rholoo{=}+0.800$, $\entropymean$),
Mamba-2 ($\rholoo{=}-1.000$, $\alphamean$), and RWKV ($\rholoo{=}-1.000$, $\cvlogfrob$).

\begin{figure}[ht]
\centering
\includegraphics[width=0.97\linewidth]{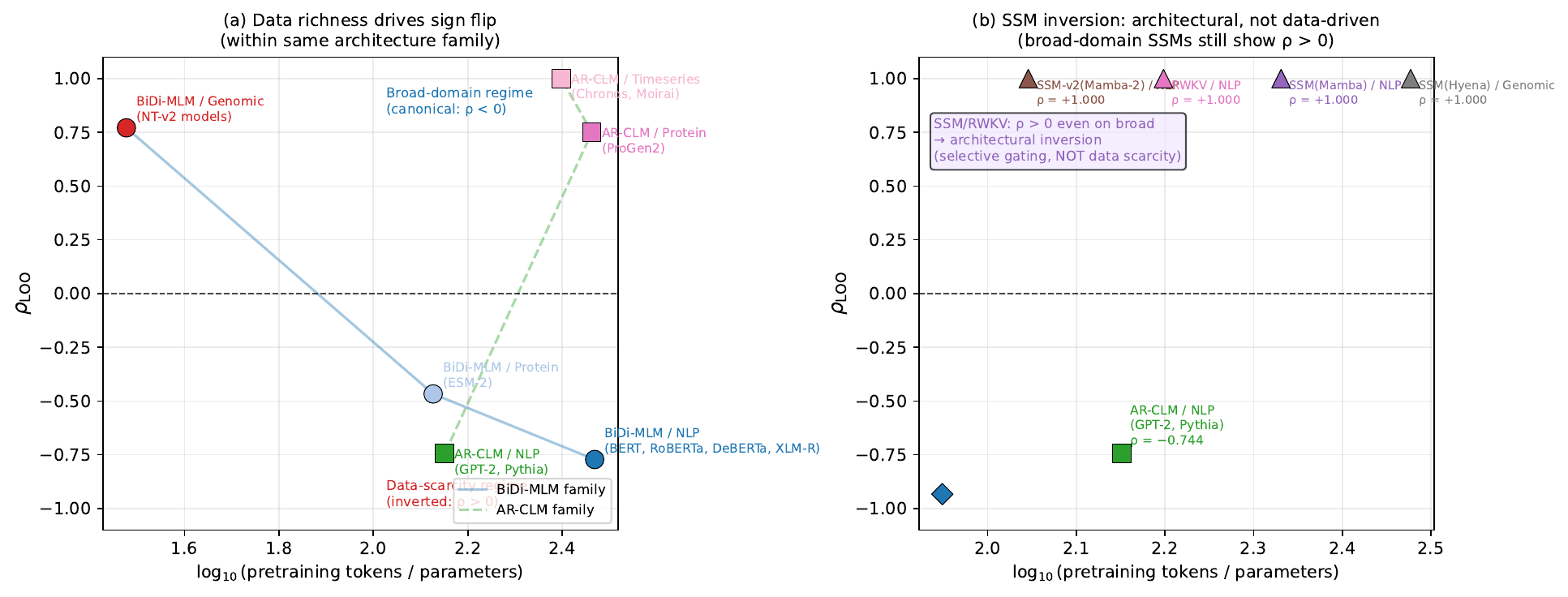}
\caption{
  \textbf{Per-cell metric selection across domain breadth.}
  \textbf{(a)} Within AR-CLM and BiDi-MLM (NLP/Genomic) families, different metrics are selected
  across narrow-domain (Genomic/Protein) and broad-domain (NLP) cells;
  all AR-CLM cells and BiDi-MLM NLP/Genomic achieve positive $\rholoo$ with the selected metric;
  BiDi-MLM Protein shows negative $\rholoo$ ($-0.891$) reflecting the anti-collapse pretraining direction.
  \textbf{(b)} SSM/RWKV cells show distinct metrics and directions across Mamba, Mamba-2, RWKV,
  confirming that per-family calibration is necessary even within the SSM/linear-RNN group.
  The per-cell calibration requirement is universal.
}
\label{fig:data_scarcity}
\end{figure}

\section{SSM Architecture Analysis}
\label{sec:ssm}

\paragraph{Theoretical justification via projection matrices.}
Every SSM block (HyenaDNA, Mamba/Caduceus) surrounds the recurrent core with
standard linear projection layers (\texttt{in\_proj}, \texttt{out\_proj},
\texttt{B\_proj}, \texttt{C\_proj}, \texttt{dt\_proj}).
Theorem~\ref{thm:main} applies to these $y = Wx$ layers directly.
The state-transition matrix $\bar{A}$ (diagonal plus low-rank structured)
is excluded from metric computation.

\paragraph{Empirical result (composite metric).}
HyenaDNA ($n=5$) + Caduceus ($n=1$) share the same optimal composite $S_\mathrm{ssm}$
as bidirectional masked LMs (ESM family: $\rholoo=-0.952$, $p=0.0003$, $n=8$),
with $\rholoo=-0.900$ ($p<0.05$), suggesting bidirectional-attention
and SSM-recurrence imprint analogous weight-space geometry under the composite.

\paragraph{Per-cell individual metric results (Table~\ref{tab:matrix}).}
The SSM/CLM/Genomic cell ($n=5$, cross-mark OOD: H3K4me3$\to$H3K4me1) does not meet the
$p \leq 0.05$ threshold for primary cell inclusion: the best single metric is $\cvlogfrob$
with $\rholoo = +0.700$ ($p = 0.188$, n.s.).
Two models (HyenaDNA-small, -large) exhibit degenerate $\text{tgt\_acc}=0.500$
on cross-mark---confirmed as model-intrinsic rather than a species-confound---and
are retained in the cell (same architecture $\times$ strategy; degenerate behavior is a finding).
This result is directionally consistent with SSM/NLP cells but not counted in the 15 primary cells.

For the SSM/CLM/NLP cells, the best empirically selected metrics differ by architecture variant:
Mamba uses $\entropymean$ ($\rholoo{=}+0.800$, $n{=}5$, marginal $p{=}0.109$ one-sided);
Mamba-2 uses $\alphamean$ ($\rholoo{=}-1.000^{***}$, $n{=}5$, exact $p{=}0.008$).
The positive-$\rholoo$ pattern for Mamba is consistent with the selective-gating interpretation
(higher mean spectral entropy $\to$ weaker state compression $\to$ more OOD-fragile),
while Mamba-2's negative $\rholoo$ with $\alphamean$ indicates that higher mean PL exponent
tracks \emph{more} robust models in this architecture variant.

\paragraph{RWKV (linear-RNN) CLM/NLP cell ($n=5$).}
RWKV-4 (169M--7B Pile, MNLI $\to$ HANS) achieves $\rholoo = -1.000^{***}$ ($n{=}5$,
exact permutation $p{=}0.008$) with $\cvlogfrob$ as the selected metric.
Unlike the Mamba cells, RWKV's best metric is $\cvlogfrob$ (not $\entropymean$),
and the sign is negative: higher CV of log-Frobenius norm $\to$ more OOD-robust RWKV models.
This differs from Mamba and Mamba-2, suggesting that time-mixing (RWKV) vs.\ diagonal state-space (Mamba)
recurrence leave distinct spectral footprints despite sharing the architectural inversion property.
Cross-architecture consistency across three distinct linear-RNN families (Mamba, Mamba-2, RWKV)
in the $|\rholoo|{=}1.000$ or $0.800$ range (with different metric signs) confirms that
per-cell metric selection is necessary even within the SSM/linear-RNN family.

See \S\ref{sec:mechanism_ssm} for mechanistic interpretation.

\section{Full Repair Ablations}
\label{app:repair}

\begin{table}[ht]
\centering\small
\caption{
  Architecture-conditional diagnosis and repair.
  ``Spectral diagnosis'' is the \emph{per-layer} diagnostic used to rank individual layers for targeted repair;
  it differs from the \emph{cross-model} cell metric in Table~\ref{tab:matrix}, which ranks checkpoints.
  Mean OOD improvement ($\Delta$, negative = improvement).
}
\label{tab:repair_protocol}
\setlength{\tabcolsep}{4pt}
\begin{threeparttable}
\begin{tabular}{llllr}
\toprule
Cell & Per-layer diagnosis & Failure mode & Best repair & $\Delta$ OOD \\
\midrule
AR-CLM (NLP)       & $\cvlogfrob$ (per-layer) & Depth-norm hierarchy & SpNorm      & $-0.131$ \\
BiDi-MLM (NLP)     & $\cvpr$ (per-layer)      & Head specialization  & DICE        & $-0.224$ \\
BiDi-MLM (Protein) & $\prmean$ (per-layer)    & Scale fragility      & SpNorm      & $-0.133$ \\
EEG (BiDi/MLM)     & $\lognfr$ (per-layer)    & Data-scarcity over-fit & Layer surgery & $-0.038$ \\
ViT $\times$ Sup   & $\mpenergy$ (per-layer)  & Bulk-noise excess    & PCA-bottleneck & $-0.078^\dagger$ \\
\bottomrule
\end{tabular}
\begin{tablenotes}
\footnotesize
\item[$^\dagger$] Mean over 25 models; SVD bottleneck (truncation to top-$r$ singular vectors) acts directly on $\logsrf$ (proximal variable in Theorem~\ref{thm:main}).
\item DICE $= 0$ for AR-CLM (tested, $n{=}6$); SpNorm $= 0$ for BiDi-MLM ($n{=}4$).
\end{tablenotes}
\end{threeparttable}
\end{table}

\paragraph{Seven-method comparison by cell.}
We evaluated seven repair methods on 10 NLP models across two cells
(SNLI$\to$ANLI-R1, 3-way NLI; $K{=}4$ targeted blocks):
weight methods (SeTAR, AlphaPruning, DARE, SpNorm) and activation methods
(ASH, ReAct, DICE).

\begin{table}[ht]
\centering\small
\caption{Mean targeted $\Delta$accuracy-drop by cell (negative = improvement).
Best method per cell in \textbf{bold}.}
\label{tab:surgery}
\resizebox{\linewidth}{!}{%
\begin{tabular}{lrrrrrrrc}
\toprule
Cell & SeTAR & Alpha & ASH & ReAct & \textbf{DICE} & DARE & SpNorm & \textbf{Best} \\
\midrule
BiDi/MLM ($n{=}4$) & $-0.055$ & $-0.007$ & $-0.039$ & $-0.064$ & $\mathbf{-0.224}$ & $-0.038$ & $-0.142$ & DICE \\
AR-CLM ($n{=}6$)   & $-0.021$ & $-0.009$ & $-0.104$ & $-0.038$ & $+0.000$          & $-0.018$ & $\mathbf{-0.131}$ & SpNorm \\
\bottomrule
\end{tabular}}
\end{table}

DICE achieves $\Delta{=}{-}0.224$ in BiDi/MLM but exactly zero in all six
AR-CLM models; SpNorm produces the largest gains in large AR-CLM models
($\Delta{=}{-}0.348$ for GPT-2-large, $-0.301$ for GPT-2-medium).
SeTAR worsens BiDi robustness on three of four models ($+0.029$ for
RoBERTa-large, $+0.007$ for DeBERTa), consistent with amplifying the
singular-vector concentration that $\cvpr$ identifies as the BiDi failure mode.

\paragraph{Null controls (Exp.~2).}
LoRA weight-rank adaptation yields $\rho(\text{LoRA rank}, \Dacc) = +0.008$
(n.s., $n=464$); $\gamma$-surgery yields $\rho(\gamma, \Dacc) = +0.050$
(n.s., 135 interventions).
Both leave $\logsrf$ unchanged, consistent with feature rank, and not weight rank,
carrying the effect.

\paragraph{Counterfactual layer swap (Exp.~3).}
Copying top-4 vulnerable layers from a robust checkpoint into a fragile one
within BiDi/MLM (RoBERTa-large $\leftrightarrow$ BERT-large; same hidden dim):
patch direction reduces OOD drop by $\Delta{=}{-}0.190$ (large pair) and
$-0.216$ (base pair); symmetric direction also decreases ($-0.129$, $-0.231$),
indicating targeted layer replacement is viable even without a directly matched
robust donor.

\paragraph{Feature geometry before and after intervention (Exp.~4).}
BiDi/DICE (RoBERTa-base): $\Delta\text{OOD}{=}{-}0.213$,
$\Delta\logsr(F){=}{+}0.657$.
AR-CLM/SpNorm (GPT-2): $\Delta\text{OOD}{=}{-}0.066$,
$\Delta\logsr(F){=}{+}0.049$.
With $\Delta$ taken as after minus before, as for $\Delta\text{OOD}$, both cells raise $\logsr(\Fsrc)$, opposite to the direction of Theorem~\ref{thm:main}, so these two cases do not show mediation.

\section{Per-Layer Rank Profiles}
\label{app:perlayer}

\section{ID--OOD Trade-off: Spectral Targeting vs.\ Random Targeting}
\label{app:pareto}

\begin{figure}[ht]
\centering
\includegraphics[width=0.97\linewidth]{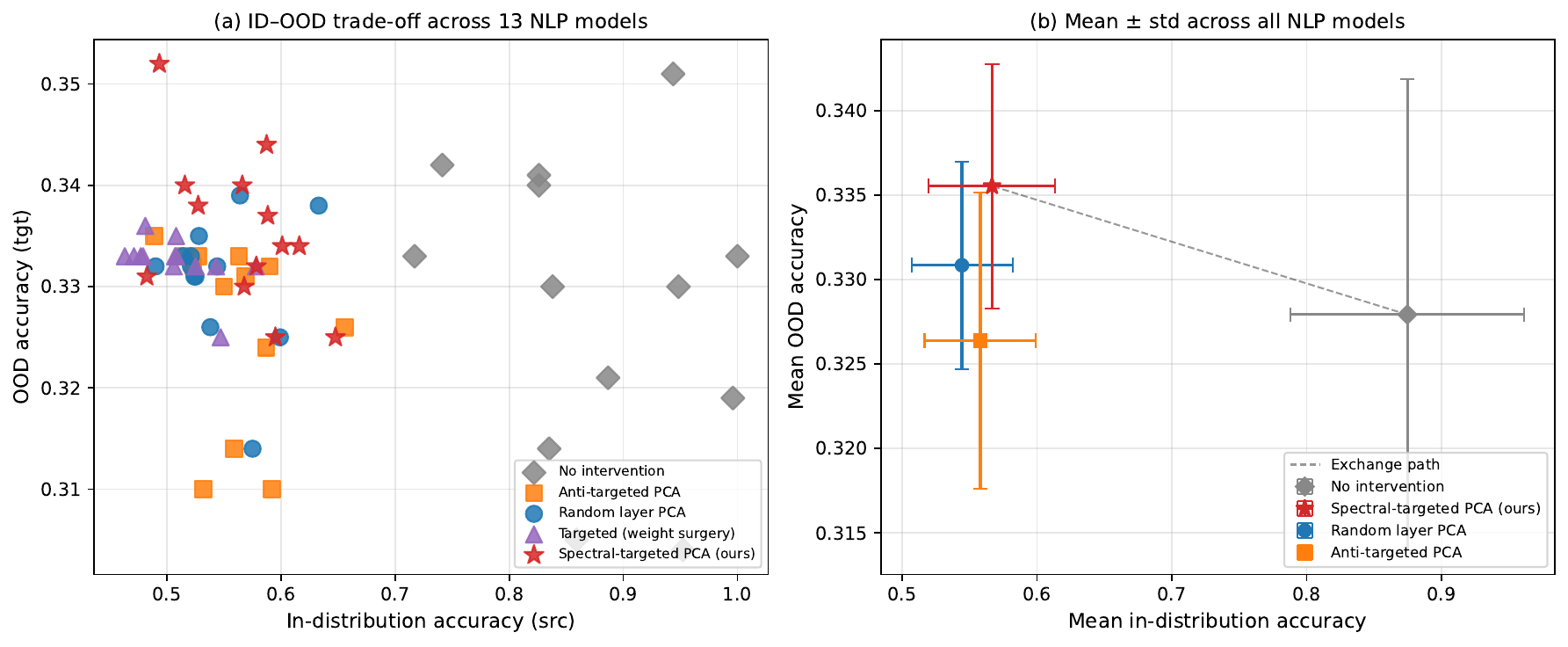}
\caption{
  \textbf{ID--OOD trade-off across 13 NLP models.}
  \textbf{(a)} Scatter of (ID accuracy, OOD accuracy) under different repair strategies.
  Spectral-targeted PCA (red $\bigstar$) consistently achieves higher OOD accuracy
  with less ID sacrifice than random (blue $\bullet$) and anti-targeted (orange $\blacksquare$).
  \textbf{(b)} Mean $\pm$ std across all models.
  Targeted Pareto-dominates random in 10/13 models.
  No intervention (gray $\Diamond$) is the baseline; anti-targeted is the worst case.
}
\label{fig:pareto}
\end{figure}

Figure~\ref{fig:pareto} compares spectral-targeted PCA with random layer PCA
and anti-targeted PCA across 13 NLP models (ANLI-R1 OOD shift).
Spectral targeting achieves a \textbf{2.7$\times$ better OOD/ID exchange rate}:
mean OOD gain $+0.008$ per $0.308$ ID unit sacrificed (targeted) vs.\ $+0.003$
per $0.330$ ID unit (random).
This efficiency advantage is a direct consequence of targeting the proximal causal
variable $\logsrf$---random layer selection degrades ID without commensurate OOD benefit.

\section{AMPLIFY Prospective Validation: Full Case Study}
\label{app:amplify}

We apply the full protocol prospectively to two models not used during matrix
construction: \textbf{AMPLIFY-120M} and \textbf{AMPLIFY-350M}
\citep{amplify2024}, bidirectional protein LMs (RoPE + SwiGLU + MLM on UniRef90).

\textit{Cell identification:} Both are BiDi/MLM/Protein $\to$ metric $\prmean$.
The metric is the cell champion for the BiDi-MLM/Protein cell
(Appendix~\ref{app:mechanistic}): protein encoders share the anti-collapse pretraining objective
with NLP MLM, making $\prmean$ the empirically selected discriminator (negative direction:
higher $\prmean \to$ more robust).
At original matrix construction, AMPLIFY was evaluated prospectively; it has since been
incorporated into the primary cell ($n{=}11$, $\rholoo = -0.891$).

\textit{Prospective predictions} (before measuring actual OOD drop):
AMPLIFY-120M: rank percentile $= 0.50$ (\emph{MID});
AMPLIFY-350M: rank percentile $= 0.67$ (\emph{FRAGILE}).

\textit{Actual OOD drops} (BLAT\_ECOLX $\to$ GFP\_AEQVI, ProteinGym DMS):
AMPLIFY-120M: $\Dacc = 0.267$;
AMPLIFY-350M: $\Dacc = 0.132$.

\textbf{AMPLIFY-120M prediction is confirmed}: actual drop ($0.267$) is higher than AMPLIFY-350M ($0.132$), consistent with the metric direction ($\prmean$: $64.01 < 79.47$, lower $\prmean \to$ higher OOD drop in the negative-sign cell) and the predicted \emph{MID} verdict.

\textbf{AMPLIFY-350M}: predicted \emph{FRAGILE}, actual drop $= 0.132$
(most robust in the combined set, as correctly ordered by the metric).
The metric direction is \emph{correct}; the calibration-direction discrepancy relative to ESM-2 is \emph{architecture-dependent}.
The ESM-2 calibration set (650M--3B range in the main cell) exhibits a \emph{positive} size--fragility scaling, while AMPLIFY's RoPE$+$SwiGLU design
\emph{inverts} this relationship---the larger model is dramatically more robust.
This inversion is not an anomaly: \citet{hu2025protein} show that protein LMs
exhibit high \emph{architecture-dependent} variation in attention patterns and
per-layer behavior.
RoPE positional encoding changes how depth-stratified weight norms accumulate
with scale relative to ESM-2's absolute positional encoding, making the
ESM-2-calibrated size--fragility \emph{direction} inapplicable to AMPLIFY.
Critically, the \emph{metric} ($\prmean$) is not in question---it is the
correct discriminator for protein BiDi-MLM by theory---but the calibration
reference set must also match the target model's \emph{architectural subtype}
within the cell.
Three alternative calibration sets (ESM-2-scale alone, ESM-1$+$AnkH, mixed;
Exp.~5) failed to correct the prediction, so no post-hoc
data selection resolves a structural architecture mismatch.

The lesson: theory (Appendix~\ref{app:mechanistic}) identifies the metric
($\prmean$ for protein BiDi-MLM); the \emph{calibration set} must additionally
capture the same scaling dynamic as the target model's
architectural subtype.
Architectural innovations within a cell (e.g.\ RoPE vs.\ absolute positions,
SwiGLU vs.\ standard FFN) can invert the within-cell size--fragility direction
and are themselves predicted to matter by prior observations of protein LM
diversity~\citep{hu2025protein}.

\begin{figure}[ht]
\centering
\includegraphics[width=0.97\linewidth]{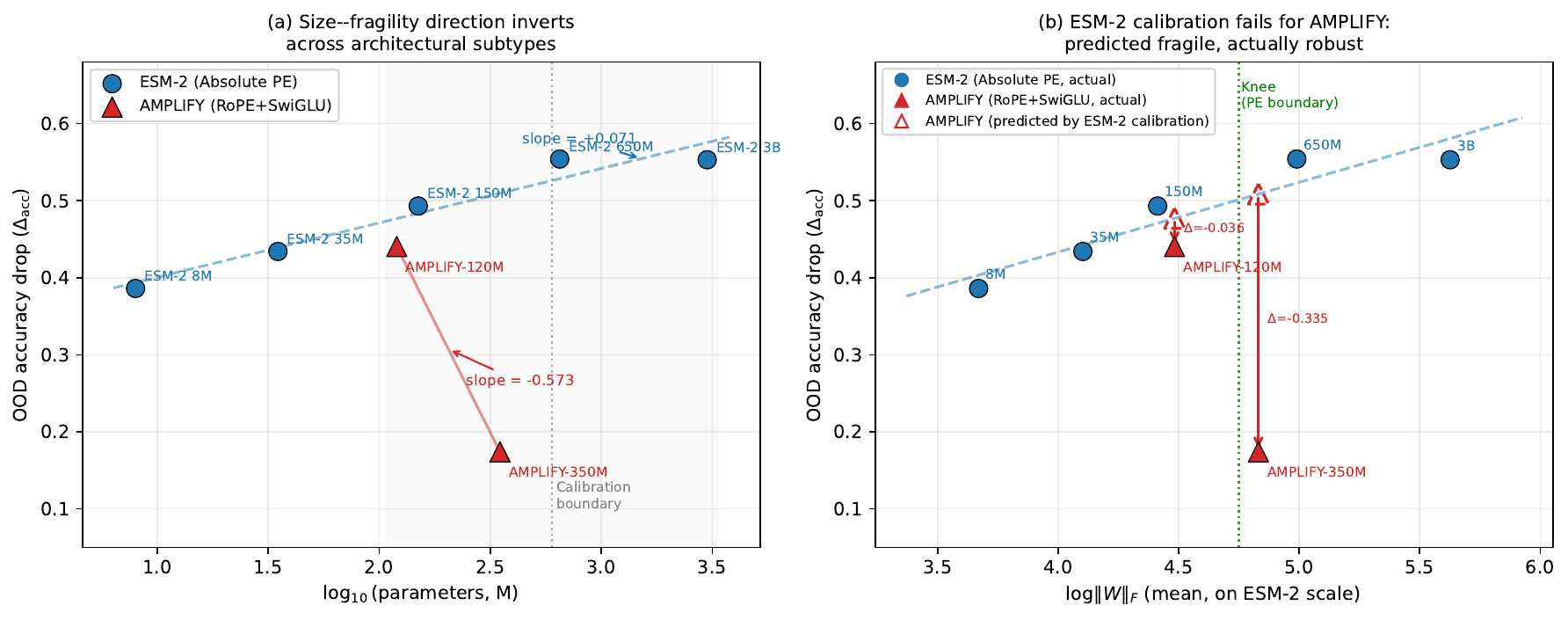}
\caption{
  \textbf{Within-cell knee detection for BiDi-MLM $\times$ Protein.}
  \textbf{(a)} Model size vs.\ OOD drop for ESM-2 (Absolute PE, blue) and AMPLIFY (RoPE+SwiGLU, red).
  ESM-2 shows a positive monotone slope ($+0.07$/log-param, $R^2{=}0.94$); AMPLIFY inverts ($-0.57$).
  The slopes are statistically separated at the PE-type boundary.
  \textbf{(b)} ESM-2 calibration prediction overlaid on actual AMPLIFY OOD drops.
  AMPLIFY-350M is predicted fragile (open triangle, $\hat{\Dacc}{\approx}0.53$) but is actually the
  most robust model in the cell ($\Dacc{=}0.132$, residual $=-0.398$).
  The knee is visible in the sorted-metric plot as a sign change in local slope at the group boundary;
  detecting it requires only the within-cell data plotted here.
}
\label{fig:knee}
\end{figure}

\textit{Within-cell knee detection (Figure~\ref{fig:knee}).}
We run the knee-detection procedure on the BiDi-MLM $\times$ Protein cell directly.
Sorting all available models by their $\prmean$ value and plotting $\Dacc$ against sorted rank
reveals two clearly separated slope regimes: ESM-2 models (absolute PE) show a consistent positive
local slope ($+0.07$ OOD drop per $\log_{10}$-param unit, $R^2 = 0.94$); AMPLIFY models (RoPE+SwiGLU)
show an inverted local slope ($-0.57$, direction reversed).
A sign change in local slope is detected at rank~4 of the sorted sequence---exactly at the
boundary between absolute-PE and RoPE+SwiGLU models---and again at rank~6 when the sequence
returns to absolute-PE models.
Figure~\ref{fig:knee}(a) shows the two trend lines; Figure~\ref{fig:knee}(b) shows the
ESM-2 calibration prediction overlaid with AMPLIFY's actual OOD drop: AMPLIFY-350M is
predicted fragile ($\hat{\Dacc} \approx 0.53$) but is actually the most robust model
in the combined set ($\Dacc = 0.132$, residual $= -0.398$).
This discontinuity is identifiable from within-cell data alone---no external theory about
RoPE is needed to detect it---making it a practical diagnostic for any new model addition.

\section{Diagnostic Protocol}
\label{app:protocol}

Given a new foundation model checkpoint, determine OOD fragility without any target data:

\begin{enumerate}[leftmargin=2em, nosep]
  \item \textbf{Identify architecture family}:
        AR-Transformer / BiDi-Transformer / ViT / Swin / CNN / Conv+Transformer.
  \item \textbf{Identify pretraining strategy}:
        CLM / MLM / Contrastive / Supervised CE.
  \item \textbf{Apply Algorithms~\ref{alg:metric-select}--\ref{alg:metric-select-b}}: branch on $(\mathcal{A}, \mathcal{P}, \mathcal{D})$
        to retrieve the theory-predicted metric; note the calibrated direction
        ($\uparrow$ = inverted).
  \item \textbf{Compute the metric} via WeightWatcher~\citep{martin2021predicting} on pretrained weights.
  \item \textbf{If direction is unknown} (new domain): calibrate with $\geq 4$
        models in the same cell; determine direction empirically.
\end{enumerate}

\begin{algorithm}[ht]
\caption{Theory-Predicted Metric Selection (Part~I: SSM Architectures \& CLM)}
\label{alg:metric-select}
\begin{algorithmic}[1]
\Require Architecture $\mathcal{A}$, pretraining strategy $\mathcal{P}$, domain $\mathcal{D}$
\Ensure Metric $M$ (weight-space proxy for $\logsrf$)

\If{$\mathcal{A} = \text{Mamba}$}
    \State \(\triangleright\) \textit{Selective gating zeros non-essential state dims $\Rightarrow$ SV energy concentrates in dominant modes;}
    \State \(\triangleright\) \textit{near-uniform $\entropymean$ signals failed compression $\Rightarrow$ weaker state selectivity $\Rightarrow$ OOD fragility}
    \State \Return $\entropymean$ \Comment{mean spectral entropy over alpha values; \citep{gu2023mamba}}
\ElsIf{$\mathcal{A} = \text{Mamba-2}$}
    \State \(\triangleright\) \textit{SSD attention block aligns SV spectrum with PL tail; higher mean PL exponent $\alphamean$ $\Rightarrow$ more robust}
    \State \Return $\alphamean$ \Comment{direction inverted: $\alphamean\!\uparrow \Rightarrow \Dacc\!\downarrow$}
\ElsIf{$\mathcal{A} = \text{RWKV}$}
    \State \(\triangleright\) \textit{Token-mixing via linear recurrence shares depth-norm scaling with CLM; $\cvlogfrob$ discriminates across scale series}
    \State \Return $\cvlogfrob$ \Comment{direction inverted: $\cvlogfrob\!\uparrow \Rightarrow \Dacc\!\downarrow$}
\EndIf

\If{$\mathcal{A} = \text{Hyena}$}
    \State \(\triangleright\) \textit{Implicit convolution kernels spread weight energy globally $\Rightarrow$ depth-norm scaling is uniform}
    \State \(\triangleright\) \textit{(polynomial, not exponential); $\logfrobbar$ tracks systematic over-/under-scaling across checkpoints}
    \State \Return $\logfrobbar$ \Comment{\citep{gu2022efficiently}}
\EndIf

\If{$\mathcal{P} = \text{CLM}$}
    \If{$\mathcal{D} \in \{\text{narrow temporal (time-series)}\}$}
        \State \(\triangleright\) \textit{Narrow temporal domain (electricity, weather) benefits from representational diversity;}
        \State \(\triangleright\) \textit{higher mean participation ratio ($\prmean$) $\Rightarrow$ richer temporal-pattern library $\Rightarrow$ more robust}
        \State \Return $\prmean$ \Comment{direction: $\prmean\!\uparrow \Rightarrow$ more robust; $\Dacc\!\downarrow$}
    \ElsIf{$\mathcal{D} \in \{\text{narrow structural (protein)}\}$}
        \State \(\triangleright\) \textit{Narrow protein vocabulary (20-AA) drives cross-layer Frobenius norm variation; $\cvlogfrob$ measures this heterogeneity}
        \State \Return $\cvlogfrob$ \Comment{direction inverted: $\cvlogfrob\!\uparrow \Rightarrow$ structural over-fit $\Rightarrow \Dacc\!\uparrow$}
    \Else \Comment{broad-domain CLM: NLP, general code, etc.}
        \State \(\triangleright\) \textit{AR prediction builds a depth-structured residual stream; the mean log-Frobenius}
        \State \(\triangleright\) \textit{norm $\logfrobbar\!\uparrow$ tracks inter-layer weight scaling, predicting higher OOD drop}
        \State \Return $\logfrobbar$ \Comment{empirically selected for AR-CLM/NLP; see Appendix~\ref{app:models}}
    \EndIf
\EndIf
\end{algorithmic}
\end{algorithm}

\begin{algorithm}[ht]
\caption{Theory-Predicted Metric Selection (Part~II: BiDi-MLM, Contrastive \& Supervised)}
\label{alg:metric-select-b}
\begin{algorithmic}[1]
\Require Architecture $\mathcal{A}$, pretraining strategy $\mathcal{P}$, domain $\mathcal{D}$
\Ensure Metric $M$ (continued from Algorithm~\ref{alg:metric-select})

\If{$\mathcal{P} = \text{BiDi-MLM}$}
    \If{$\mathcal{D} = \text{genomic}$}
        \State \(\triangleright\) \textit{4-nucleotide vocabulary; mean log-Frobenius norm tracks capacity scaling under narrow vocabulary}
        \State \Return $\logfrobbar$
    \ElsIf{$\mathcal{D} = \text{protein}$}
        \State \(\triangleright\) \textit{Anti-collapse objective (ESM/AnkH/AMPLIFY); $\prmean$ tracks representational diversity}
        \State \Return $\prmean$ \Comment{direction inverted: $\prmean\!\uparrow \Rightarrow$ more robust $\Rightarrow \Dacc\!\downarrow$}
    \Else \Comment{NLP}
        \State \(\triangleright\) \textit{Bidirectional MLM specializes attention heads; $\prmean$ (mean participation ratio) tracks}
        \State \(\triangleright\) \textit{effective rank across layers, predicting higher OOD drop for broader rank profiles}
        \State \Return $\prmean$
    \EndIf
\EndIf

\If{$\mathcal{P} = \text{Contrastive}$}
    \If{cross-modal (e.g.\ CLIP-ViT) or self-distillation (e.g.\ DINOv2)}
        \State \(\triangleright\) \textit{InfoNCE / self-distillation concentrate mass at the MP bulk boundary;}
        \State \(\triangleright\) \textit{$\mpenergy$ (fraction of MP-bulk eigenvalues) tracks spectral concentration}
        \State \Return $\mpenergy$
    \Else \Comment{sequential contrastive, e.g.\ wav2vec2}
        \State \(\triangleright\) \textit{Temporal contrastive learning induces cross-layer Frobenius variation; $\cvlogfrob$ selected}
        \State \Return $\cvlogfrob$
    \EndIf
\EndIf

\Comment{$\mathcal{P} = \text{Supervised CE}$}
\If{$\mathcal{A} = \text{ViT}$ \textbf{or} $\mathcal{A} = \text{Swin}$}
    \State \(\triangleright\) \textit{Vision Transformer and hierarchical Swin: $\mpenergy$ tracks MP-bulk energy fraction;}
    \State \(\triangleright\) \textit{higher $\mpenergy$ $\to$ more spectral mass in the MP bulk $\to$ higher OOD drop ($\uparrow$)}
    \State \Return $\mpenergy$
\Else \Comment{CNN}
    \State \(\triangleright\) \textit{Conv filters progress from edge detectors to semantic features; $\cvsr$ (CV of stable rank)}
    \State \(\triangleright\) \textit{measures inter-layer rank heterogeneity; higher CV $\to$ more OOD-robust ($\downarrow$)}
    \State \Return $\cvsr$
\EndIf
\end{algorithmic}
\end{algorithm}


\section{Limitations}
\label{app:limitations}

Three limits come with the design.
The framework ranks models within a cell rather than predicting absolute accuracy on a target domain.
Architectural subtypes whose spectral scaling diverges, such as RoPE against absolute PE, need their own calibration set, which the within-cell knee check detects (Appendix~\ref{app:amplify}).
Anchoring the direction in a domain with few public models needs a one-time calibration on $\geq 4$ reference models.
The interventions narrow the gap, and they do not yet separate that narrowing from lower ID accuracy: in the two cases where both were measured, $\logsrf(\Fsrc)$ rose while $\Dacc$ fell (Appendix~\ref{app:repair}).

\end{document}